\documentclass[a4paper, 11pt, twoside]{article}
\usepackage{fullpage}

\usepackage{amsmath,amsfonts,amssymb,amsthm,array}
\usepackage{bm}
\usepackage{wrapfig}
\usepackage[colorinlistoftodos,bordercolor=orange,backgroundcolor=orange!20,linecolor=orange,textsize=scriptsize]{todonotes}
\usepackage{enumitem}
\usepackage[algo2e,ruled,linesnumbered]{algorithm2e}
\allowdisplaybreaks
\usepackage{makecell}
\usepackage{caption}
\usepackage{multirow}

\usepackage{colortbl}
\definecolor{bgcolor}{rgb}{0.8,1,1}
\definecolor{bgcolor2}{rgb}{0.8,1,0.8}
\usepackage{threeparttable}

\newcommand{\myred}[1]{{\color{red}#1}}

\usepackage{tcolorbox}
\usepackage{pifont}
\definecolor{mydarkgreen}{RGB}{39,130,67}
\definecolor{mydarkred}{RGB}{192,47,25}
\newcommand{\green}{\color{mydarkgreen}}
\newcommand{\red}{\color{mydarkred}}
\newcommand{\cmark}{{\green\ding{51}}}
\newcommand{\xmark}{{\red\ding{55}}}
\usepackage{subcaption}

\newcommand{\EndProof}{\begin{flushright}$\square$\end{flushright}}

\newcommand*\EE[1]{\ensuremath{\mathbb{E}[||#1||^2]}}

\newcommand{\calO}[0]{\mathcal{O}}

\usepackage{booktabs}

\newcommand{\Exp}{\mathbb{E}}

\newcommand{\R}{\mathbb{R}}

\newcommand{\ab}[1]{{\color{black} #1}} 

\theoremstyle{plain}

\theoremstyle{remark}

\usepackage{mdframed} 
\usepackage{thmtools}

\usepackage{caption}
\usepackage{multirow}
\usepackage{colortbl}
\definecolor{bgcolor}{rgb}{0.8,1,1}
\definecolor{bgcolor2}{rgb}{0.8,1,0.8}
\definecolor{niceblue}{rgb}{0.0,0.19,0.56}

\usepackage{hyperref}
\hypersetup{colorlinks,linkcolor={blue},citecolor={niceblue},urlcolor={blue}}

\usepackage{natbib}
\usepackage{sidecap}

\definecolor{shadecolor}{gray}{0.9}
\declaretheoremstyle[
headfont=\normalfont\bfseries,
notefont=\mdseries, notebraces={(}{)},
bodyfont=\normalfont,
postheadspace=0.5em,
spaceabove=1pt,
mdframed={
  skipabove=8pt,
  skipbelow=8pt,
  hidealllines=true,
  backgroundcolor={shadecolor},
  innerleftmargin=4pt,
  innerrightmargin=4pt}
]{shaded}

\declaretheorem[style=shaded]{theorem}

\declaretheorem[style=shaded]{assumption}
\declaretheorem[style=shaded]{corollary}

\declaretheorem[style=shaded]{lemma}

\usepackage{xspace}

\usepackage[colorinlistoftodos,bordercolor=orange,backgroundcolor=orange!20,linecolor=orange,textsize=scriptsize]{todonotes}

\title{Random-reshuffled SARAH does not need full gradient computations}

\begin{document}

\title{\textbf{Random-reshuffled SARAH does not need full gradient computations}
}

\author{Aleksandr Beznosikov $^{1,2}$
\quad  Martin Tak\'a\v{c} $^{2}$
}
\date{$^1$ Moscow Institute of Physics and Technology, Moscow, Russian Federation\\
$^2$ Mohamed bin Zayed University of Artificial Intelligence, Abu Dhabi, UAE
}

\maketitle

\begin{abstract}
The StochAstic Recursive grAdient algoritHm (SARAH) algorithm is a variance reduced variant of the Stochastic Gradient Descent (SGD) algorithm that needs a gradient of the objective function from time to time. In this paper, we remove the necessity of a full gradient computation. This is achieved by using a randomized reshuffling strategy and aggregating stochastic gradients obtained in each epoch. The aggregated stochastic gradients serve as an estimate of a full gradient in the SARAH algorithm.
We provide a theoretical analysis of the proposed approach and conclude the paper with numerical experiments that demonstrate the efficiency of this approach.
\end{abstract}

\section{Introduction}

In this paper, we address the problem of minimizing a finite-sum problem of the form
\begin{equation}
    \label{prob}
    \min_{w \in \R^d} \left\{P(w)
     := \frac1n \sum_{i=1}^n f_i(w) \right\}, 
\end{equation}
where $\forall i \in [n] := \{1,2,\dots,n\}$ the $f_i$
is a convex function.
We will further assume that $w^* = \arg\min P(w)$ exists.

Problems of this form are very common in e.g., supervised learning \citep{shalev2014understanding}.
Let a training dataset
consists of $n$ pairs, i.e., $\{(x_i, y_i)\}_{i=1}^n$, where $x_i \in \R^d$ is a feature vector for a datapoint $i$ and $y_i$
is the corresponding label.
Then for example, the least squares regression problem corresponds to \eqref{prob}
with $f_i(w) = \frac12 (x_i^T w - y_i)^2$.
If $y_i\in\{-1,1\}$ would indicate a class, then a logistic regression is obtained by choosing
$f_i(w) = \log(1+\exp(-y_i x_i^T w))$.

Recently, many algorithms have been proposed for solving \eqref{prob}. 
In this paper, we are interested in a subclass of these algorithms that fall into a stochastic gradient descent (SGD) framework originating from the work of Robbins and Monro in '50s 
\citep{robbins1951stochastic}.
Let $v_s$ will be some sort of (possibly stochastic and very rough) approximation of $\nabla P(w_s)$, then many SGD type algorithms update the $w$ 
as follows:
\begin{equation}
w_{s+1} = w_s - \eta_s v_s, \label{eq:SGDstep}
\end{equation}
where $\eta_s > 0$ is a predefined step-size.
The classical SGD defines $v_s = \nabla f_i(w_s)$, 
where $i\in [n]$ is chosen randomly 
\citep{shalev2011pegasos}
or its mini-batch version \citep{takac2013mini}, where   $v_s = \frac1{|S|} \sum_{i \in S} \nabla f_i(w_s)$,
with $S\subset[n]$.
Even with an unbiased gradient estimates of SGD, where $\Exp[v_s | w_s] = \nabla P(w_s)$, the variance of $v_s$ is the main source of slower convergence \citep{bottou2018optimization,nguyen2019new,khaled2020better}.

\subsection{Brief Literature Review}\label{sec:rules}
Recently, many  variance-reduced variants of SGD have been proposed, including SAG/SAGA \citep{schmidt2017minimizing,defazio2014saga,qian2019saga}, SVRG \citep{johnson2013accelerating,allen2016improved,yang2021accelerating},
MISO \citep{mairal2015incremental},
SARAH \citep{nguyen2017sarah,nguyen2021inexact,nguyen2017stochastic,hu2019efficient}, 
SPIDER \citep{fang2018spider}, STORM \citep{cutkosky2019momentum}, PAGE~\citep{li2021page}, and many others.
Generally speaking, the variance-reduced variants of SGD still aim to sample $\calO(1)$ functions and use their gradients to update $v_s$.
For example, SVRG \citep{johnson2013accelerating} will fix a point $\tilde w$, in which a full gradient 
$\nabla P(\tilde w)$
is computed and subsequently stochastic gradient is defined as
$v_s = \nabla f_i(w_s) - \nabla f_i(\tilde w) + \nabla P(\tilde w)$, where $i\in [n]$ is picked at random.

\paragraph{SARAH Algorithm.}
The SARAH algorithm \citep{nguyen2017sarah}, on the other hand, updates $v_s$
recursively. It starts with a full gradient computation $v_0 = \nabla P(w_0)$, then taking a step  \eqref{eq:SGDstep}
and updating the 
gradient estimate recursively as
$
v_{s} = \nabla f_i(w_s) - \nabla f_i(w_{s-1}) + v_{s-1}$.
For a smooth and strongly convex problem, the procedure highlighted above converges, but not to the optimal solution $w^*$.
Therefore, similarly to SVRG, the process is restarted after i) a predefined number of iterations, ii) randomly
\citep{li2021zerosarah,li2020convergence,li2021page},
or iii) decided in run-time by computing the ratio 
$\|v_s\| / \|v_0\|$ (SARAH+ \citep{nguyen2017sarah}),
and a new full gradient estimate has to be computed.
To elevate this issue, e.g., in \citep{nguyen2021inexact},
they proposed inexact SARAH (iSARAH), where the full gradient estimate is replaced by a mini-batch gradient estimate
$v_0 = \frac1{|S|}\sum_{i\in S} f_i(w_0)$,
where $S\subset[n]$.
To find a point $
\hat w$ such that
$\EE{\nabla P(\hat w)} \leq \epsilon$, 
the mini-batch size has to be chosen as $|S|\sim \calO(\frac1\epsilon)$, and
the step-size will be $\eta \sim \calO(\frac\epsilon L)$.

There are a few variants of SARAH that do not need any restart and full-gradient estimate.
E.g., the Hybrid Variance-Reduced variant \citep{liu2020optimal}
defines 
\begin{equation}
v_s = \beta \nabla f_i(w_s) 
+ (1-\beta) 
\left(
\nabla f_i(w_s) 
-
\nabla f_i(w_{s-1})
+ v_{s-1}
\right),
\label{eq:storm}
\end{equation}
where $\beta \in (0,1)$
is a hyperparameter.
The STORM variant
\citep{cutkosky2019momentum}
uses \eqref{eq:storm}
not with a fixed value of parameter $\beta$, but 
in STORM, the value of $\beta_s$ is diminishing to $0$. 
The ZeroSARAH \citep{li2021zerosarah}
is another variant,
where the $v_s$ is 
a combination of \eqref{eq:storm}
with SAG/SAGA.

\paragraph{Random Sampling vs. Random Reshuffle.}
All the stochastic algorithms discussed so far sample function $f_i$
randomly.
However, it is standard practice, for a finite-sum problem, not to choose functions $f_i$ randomly
with replacement, but rather make a data permutation/shuffling
and then choose the $f_i$s in a cyclic fashion \citep{bottou2009curiously,recht2013parallel}. 
\ab{
This is primarily because methods involving data shuffling are more cache-friendly, which leads to a significant acceleration in their implementation, as noted in \citep{bengio2012practical}. From learning and optimization perspectives, a cyclic pass through the data appears to be more beneficial as it ensures that each data point is utilized an equal number of times throughout the training process. Hence, shuffling strategies are frequently used for real-world practical problems, including those encountered in deep learning, as outlined in \citep{sun2020optimization}.
}

\ab{The basic shuffling approaches include}
\begin{itemize}[noitemsep,topsep=0pt]
    \item 
    {\bf Random Reshuffling (RR)} - reshuffle data before each epoch;
 \item 
  {\bf Shuffle-Once (SO)} - shuffle data only once before optimizing;
 \item 
 {\bf Incremental Gradient (IG)} - access data in a cycling fashion over the given dataset.
\end{itemize}
\ab{
In recent years, numerous studies have offered theoretical analyses of various shuffling techniques, both in the context of classical SGD \citep{ying2018stochastic,jain2019sgd,ahn2020sgd,mishchenko2020random,nguyen2021unified,koloskova2023shuffle} and in variance reduction methods \citep{vanli2016stronger,gurbuzbalaban2017convergence,mokhtari2018surpassing,sun2019general,park2020linear,ying2020variance,malinovsky2021random,NEURIPS2021_1a3650ae}.
}

\paragraph{} \hspace{-0.15cm}\ab{In Table \ref{tab:comparison0} we summarize the results from the literature on stochastic methods with independent generation of batch numbers and with shuffling techniques. In the case of independent generation, the results do not claim to be complete and are given rather for information and comparison.}

\begin{table}[h!]
\renewcommand{\arraystretch}{1.5}
    \centering
    \scriptsize
\captionof{table}{Summary of results on solving a smooth strongly convex finite-sum optimization problem. The results are given both for independent generation of batch numbers and for different shuffling techniques. In the case of independent generation, the results are given for reference and make no claim to completeness.}
    \label{tab:comparison0}   
  \begin{threeparttable}
    \begin{tabular}{ c c c c c c c }
    \toprule
    \multicolumn{1}{c}{} & \textbf{Method} & \textbf{References} & 
    \multicolumn{1}{p{8mm}}{
    \textbf{No Full Grad.?}
    }
    
     & \textbf{Memory} & 
     \multicolumn{1}{p{8mm}}{
    \textbf{Lin. Conv.?}
    }
  & \textbf{Complexity}
    \\

\midrule
     
    \multirow{9}{*}{\rotatebox[origin=c]{90}{\textbf{Independent}}} &
    GD & \citep{polyak1987introduction} & \xmark &  $\mathcal{O}\left( d \right)$ & \cmark & $\mathcal{O}\left( n\frac{L}{\mu} \log \frac{1}{\varepsilon}\right)$
    \\
    \cline{2-7}
    & AGD & \citep{nesterov2003introductory} & \xmark &  $\mathcal{O}\left( d \right)$ & \cmark & $\mathcal{O}\left( n\sqrt{\frac{L}{\mu}} \log \frac{1}{\varepsilon}\right)$
    \\
    \cline{2-7}
    & SGD &
    \begin{tabular}{@{}c@{}}
    \citep{robbins1951stochastic}
    \\
    [-2mm]
    \citep{bachmoulines2011}
    \\
    [-2mm]
    \citep{ghadimi2016mini}
    \\
    [-2mm]
    \citep{stich2019unified}
    \end{tabular}
    & \cmark &  $\mathcal{O}\left( d \right)$ & \xmark & $\mathcal{O}\left( \frac{L}{\mu} \log \frac{1}{\varepsilon} + \frac{\sigma^2_*}{\mu^2 \varepsilon}\right)$
    \\
    \cline{2-7}
    & ASGD & 
    \begin{tabular}{@{}c@{}}
    \citep{cohen2018acceleration}
    \\
    [-2mm]
    \citep{vaswani2019fast} 
    \end{tabular}
    & \cmark &  $\mathcal{O}\left( d \right)$ & \xmark & $\mathcal{O}\left( \sqrt{\frac{L}{\mu}} \log \frac{1}{\varepsilon} + \frac{\sigma^2}{\mu^2 \varepsilon} \right)$
    \\
    \cline{2-7}
    & SAGA & \citep{defazio2014saga} & \cmark &  \myred{$\mathcal{O}\left( nd \right)$} & \cmark & $\mathcal{O}\left( \left[ n + \frac{L}{\mu} \right] \log \frac{1}{\varepsilon}\right)$
    \\
    \cline{2-7}
    & SVRG & \citep{johnson2013accelerating} & \xmark &  $\mathcal{O}\left( d \right)$ & \cmark & $\mathcal{O}\left( \left[ n + \frac{L}{\mu} \right] \log \frac{1}{\varepsilon}\right)$
    \\
    \cline{2-7}
    & SARAH & \citep{nguyen2017sarah} & \xmark &  $\mathcal{O}\left( d \right)$ & \cmark & $\mathcal{O}\left( \left[ n + \frac{L}{\mu} \right] \log \frac{1}{\varepsilon}\right)$
    \\
    \cline{2-7}
    & Katyusha & \citep{allen2017katyusha} & \xmark &  $\mathcal{O}\left(d \right)$ & \cmark & $\mathcal{O}\left( \left[n + \sqrt{n\frac{L}{\mu}} \right]\log \frac{1}{\varepsilon}\right)$
    \\
    \midrule
    \multirow{10}{*}{\rotatebox[origin=c]{90}{\textbf{Shuffling}}} & 
    SGD &
    \begin{tabular}{@{}c@{}}
\citep{ying2018stochastic}
    \\
    [-2mm]
\citep{jain2019sgd,ahn2020sgd}
    \\
    [-2mm]
    \citep{ahn2020sgd}
    \\
    [-2mm]
    \citep{mishchenko2020random}
    \\
    [-2mm]
    \citep{nguyen2021unified}
    \end{tabular}
    & \cmark & $\mathcal{O}\left( d\right)$ & \xmark & $\mathcal{O}\left( \frac{L}{\mu} \log \frac{1}{\varepsilon}  + \sqrt{\frac{L \sigma^2_*}{\mu^3 n \varepsilon}}\right)$
    \\
    \cline{2-7}
    & IAG & 
    \begin{tabular}{@{}c@{}}
    \citep{vanli2016stronger}
    \\
    [-2mm]
    \citep{gurbuzbalaban2017convergence} 
    \end{tabular}
    & \cmark &  \myred{$\mathcal{O}\left( nd \right)$} & \cmark & $\mathcal{O}\left( n \frac{L}{\mu} \log \frac{1}{\varepsilon} \right)$
    \\
    \cline{2-7}
    & DIAG  & \citep{mokhtari2018surpassing} & \cmark &  \myred{$\mathcal{O}\left( nd \right)$} & \cmark & $\mathcal{O}\left( n \frac{L}{\mu} \log \frac{1}{\varepsilon} \right)$
    \\
    \cline{2-7}
    & SAGA & 
    \begin{tabular}{@{}c@{}}
    \citep{sun2019general} 
    \\
    [-2mm]
    \citep{park2020linear}
    \\
    [-2mm]
    \citep{ying2020variance} 
    \end{tabular}
    & \cmark &  \myred{$\mathcal{O}\left( nd \right)$} & \cmark & $\mathcal{O}\left( n \frac{L^2}{\mu^2} \log \frac{1}{\varepsilon} \right)$
    \\
    \cline{2-7}
    & SVRG & 
    \begin{tabular}{@{}c@{}}
    \citep{sun2019general}
    \\
    [-2mm]
    \citep{malinovsky2021random}
    \end{tabular}
    & \xmark & $\mathcal{O}\left( d \right)$ & \cmark & $\mathcal{O}\left( n \frac{L^{3/2}}{\mu^{3/2}} \log \frac{1}{\varepsilon} \right)$
    \\
    \cline{2-7}
    & AVRG & \citep{ying2020variance} & \cmark & $\mathcal{O}\left( d\right)$ & \cmark & $\mathcal{O}\left( n \frac{L^2}{\mu^2} \log \frac{1}{\varepsilon} \right)$
    \\
    \cline{2-7}
    & Prox-DFinito & \citep{NEURIPS2021_1a3650ae} & \cmark & \myred{$\mathcal{O}\left( nd \right)$} & \cmark & $\mathcal{O}\left( n \frac{L}{\mu} \log \frac{1}{\varepsilon} \right)$
    \\
    \cline{2-7}
    &\cellcolor{bgcolor2}{Algorithm 1} & \cellcolor{bgcolor2}{this paper} & \cellcolor{bgcolor2}{\cmark} & \cellcolor{bgcolor2}{$\mathcal{O}\left( d \right)$} & \cellcolor{bgcolor2}{\cmark} & \cellcolor{bgcolor2}{$\mathcal{O}\left( \left[ n \frac{L}{\mu} + n^2 \frac{\delta}{\mu} \right] \log \frac{1}{\varepsilon} \right)$}
    \\
    \cline{2-7}
    &\cellcolor{bgcolor2}{Algorithm 2} & \cellcolor{bgcolor2}{this paper} & \cellcolor{bgcolor2}{\xmark} & \cellcolor{bgcolor2}{$\mathcal{O}\left( d \right)$} & \cellcolor{bgcolor2}{\cmark} & \cellcolor{bgcolor2}{$\mathcal{O}\left( \left[ n \frac{L}{\mu} + n^2 \frac{\delta}{\mu} \right] \log \frac{1}{\varepsilon} \right)$}
    \\
    \bottomrule
    \end{tabular}   
    \begin{tablenotes}
    \item [] {\em Columns:} No Full Grad.? = whether the method computes full gradients, Memory = number of additional memory, Lin. Conv.? = whether the method converges linearly to the solution $w^*$, Complexity = oracle complexity on the terms $f_i$ to find the solution (give the best result of the works listed in References).
    \item [] {\em Notation:} $\mu$ = constant of strong monotonicity of $P$, $L$ = Lipschitz constant of $\nabla f_{i}$, $\delta=$ similarity parameter, $\sigma^2_* = \tfrac{1}{n} \sum_{i=1}^n \| \nabla f_i (w^*)\|^2$ (in \citep{mishchenko2020random}, the authors use a slightly different definition), $\sigma^2$ = upper bound for $\| \nabla f_i (w) - \nabla P(w)\|^2$, $n$ =  size of the dataset, $d$ = dimension of the problem, $\varepsilon$ = accuracy of the solution. 
\end{tablenotes}    
    \end{threeparttable}
\vspace{-0.5cm}
\end{table}
 
\subsection{Contribution}  
The main contribution of this paper is the modification of the SARAH algorithm to remove the requirement of computing a full gradient $\nabla P(w)$, while achieving a linear convergence with a fixed step-size for strongly convex objective. The crucial algorithmic modification that was needed to achieve this goal, was to replace the random selection of functions by either of the shuffle options ({\bf RR}, {\bf SO}, {\bf IG})
and designing a mechanism that can build a progressively better approximation of a full gradient $\nabla P(w_s)$
as $w_s \to w^*$.

\ab{One can note that, according to Table \ref{tab:comparison0}, our Algorithm \ref{Shuffled-SARAH} is one of the few methods with shuffling that simultaneously does not compute full gradients, does not use large amounts of additional memory, and converges linearly for smooth strongly convex finite-sum problems \eqref{prob}. Moreover, with a sufficiently small $\delta$ (see the discussion after Assumption \ref{ass} and at the end of Section \ref{theory}), Algorithm \ref{Shuffled-SARAH} has state-of-the-art guarantees of convergence among methods using the shuffling technique.}

\section{\texttt{Shuffled-SARAH}}

\subsection{Building Gradient Estimate While Optimizing}
\paragraph{An intuition.}
\ab{
By accessing data in a cyclic order, employing any of the alternative methods mentioned earlier, we can approximate the full gradient $\tilde v \approx \nabla P$. In fact, if we set the step-size $\eta_s$ in \eqref{eq:SGDstep} to zero and let $v_s$ represent a stochastic gradient $\nabla f_i$, then by averaging all the stochastic gradients in a single pass, we would obtain the exact full gradient $\nabla P(w)$. As we increase the value of $\eta$, the stochastic gradients would be computed at different points}
\begin{equation}
\label{eq:tildeVform}
    \tilde v = \frac1n\sum_{i=1}^n \nabla f_{\pi^i} (w_i),
\end{equation}  and hence we would not obtain the exact full gradient of $P(w)$ but rather just a rough estimate. But is it just the $\eta$ that affects how good the $\tilde v$ will be? Of course not, as $w_s$ is updated using \eqref{eq:SGDstep}, one can see that the radius of a set of $w$s that are used to compute gradient estimates is dependent on $v_s$. Ideally, as we will converge go $w^*$, then also $v_s \to \nabla P(w^*) = 0$ and hence $\tilde v$ will be getting closer to $\nabla P(w_s)$. 

\paragraph{Building the gradient estimate.}
Our proposed approach to eliminate the need to compute the full gradient is based on a simple recursive update.
Let us initialize $\tilde v_0 = {\bf 0}\in \R^d$.
Then while making a pass $i=\{1,2,\dots,n\}$ over the data, we will keep updating $\tilde v$ using the gradient estimates
as follows
$$
\tilde v_i = \frac{i-1}{i} \tilde v_{i-1} + \frac1i \nabla f_{\pi^i} (w_i), \quad \mbox{for}\ i\in\{1,2,\dots,n\}.
$$
It is an easy exercise to see that $\tilde v_i$ will be the average of gradients seen so far, and moreover, after $n$ updates, it will be exactly as in \eqref{eq:tildeVform}.
Let us note that making the pass over the dataset is crucial to build a good estimate of the gradient and random selection of functions would not achieve this goal.

\paragraph{The Algorithm.}
\ab{
We are now ready to explain the \texttt{Shuffled-SARAH} algorithm, as outlined in Algorithm~\ref{Shuffled-SARAH}.
The algorithm commences with the selection of an initial solution $w^{-}$, which can be randomly chosen or simply set to, for example, ${\bf 0}$. Subsequently, we define $v_0 = {\bf 0}$, which will consistently serve as an estimate of the full gradient $\nabla P$.

In line \ref{pointer}, we assign $\tilde v$ to share the same memory address as $v_0$. Essentially, this implies that during the first pass $s = 0$, $v_0$ and $\tilde v$ will always be identical, and any modification to $\tilde v$ will concurrently affect $v_0$. After executing lines \ref{vs}, \ref{tvs}, $v_s$ and $\tilde v$ will evolve into two distinct vectors.

The reason we have included the term {\it in place} in line \ref{inplace} is solely to ensure that for $s=0$, both $v_0$ and $\tilde v$ will remain the same.
}


The random permutation in line \ref{randomPermutation}
could have one of the three options mentioned in Section~\ref{sec:rules}.
For {\bf RR}, we will permute the $[n]$ each time, 
for {\bf SO} we will only shuffle once for $s=0$ and define $\pi_{s} = \pi_0$ for any $s>0$.
In {\bf IG} option we have $\pi_s = (1,2,\dots,n)$
$\forall s$.
 




\begin{algorithm2e}[t]
\caption{\texttt{Shuffled-SARAH}}
\label{Shuffled-SARAH}
 \SetAlgoLined
  {\bf Input:} $0<\eta $ step-size
  
  choose $w^{-} \in \R^d$

  $w = w^{-}$

  $v_0 = {\bf 0}\in \R^d$ 
  
  $\tilde v = \&(v_0)$  \CommentSty{\qquad // $\tilde v$ will point to $v_0$} 
  \label{pointer}

  $\Delta = {\bf 0} \in \R^d$
  
  \For{$s = 0, 1, 2, \dots$}{
      define $w_s := w$
      
      $w^- = w$
      
      $w = w - \eta v_s$
      
      obtain permutation $\pi_s = (\pi_s^1, \dots, \pi_s^n)$ of $[n]$ by some rule 
      \label{randomPermutation}
      
        \For{$i=1,2,\dots,n$}{ 
            $\tilde v = \frac{i-1}{i} \tilde v + \frac1{i} \nabla f_{\pi_s^i} (w) $  
            \label{inplace}
            
            $\Delta = \Delta + \nabla  f_{\pi_s^i} (w)
                             - \nabla f_{\pi_s^i} (w^{-}) $ 
                             
             $w^{-} = w$
            
            $w = w - \eta (v_s + \Delta)$

        }

      $v_{s+1} = \tilde v$
      \label{vs}
      
      $\tilde v = {\bf 0}\in \R^d$
      \label{tvs}
      
      $\Delta = {\bf 0} \in \R^d$
      
    } 
  
    {\bf Return: } $w$
\end{algorithm2e}

\section{Theoretical Analysis} \label{theory}

\ab{Before presenting our theoretical results we introduce some notations and assumptions.
}

We use $\langle x,y \rangle := \sum_{i=1}^d x_i y_i$ to define standard inner product of $x,y\in\R^d$. It induces $\ell_2$-norm in $\R^d$ in the following way $\|x\| := \sqrt{\langle x, x \rangle}$.

\begin{assumption}\label{ass}  For problem \eqref{prob} the following hold:
 \begin{itemize}[noitemsep,topsep=0pt]
      \item[(i)] Each $f_i:\mathbb{R}^{d}\to \mathbb{R}$ is convex and twice differentiable, with $L$-Lipschitz gradient: 
      \begin{equation*}
    \|\nabla f_i(w_1) - \nabla f_i(w_2)\| \leq L\|w_1-w_2\|,
   \end{equation*}
for all   $w_1, w_2\in \mathbb{R}^{d}$;
     \item[(ii)]  $P(w)$ is a $\mu$-strongly convex function with minimizer $x^*$ and optimal value $P^*$;
     \item[(iii)] Each $f_i$ is $\delta$-similar with $P$, i.e. for all   $w \in \mathbb{R}^{d}$ it holds that
     \begin{equation*}
    \|\nabla^2 f_i(w) - \nabla^2 P(w)\| \leq \delta/2,
   \end{equation*}
   where $\| \cdot \|$ is the operator norm between matrices.
 \end{itemize}
\end{assumption}
\ab{ 
The last assumption pertains to the Hessian similarity of $\{f_i\}_{i=1}^n$.
This effect is prominently seen when the data with $n\cdot b$ samples is  randomly partitioned  across $n$
batches 
$ \{ \{z_j^i\}_{j=1}^b\}_{i=1}^{n} $
and $f_i$ is defined as
$f_i(w) = \frac{1}{b} \sum_{j=1}^b \ell (w, z_j^i)$, where $\ell$ is a $L$-smooth and convex loss function.
One can see that
$P(w) =\frac1n \sum_{i=1}^n f_i(w) =  \frac{1}{nb} \sum_{i=1}^n\sum_{j=1}^{b} \ell (w, z_j^i)$
forms the total loss function 
over the whole dataset.

Subsequently, one can use Hoeffding’s inequality for matrices \citep{tropp2012user} and obtain that with probability $ 1- p$ it holds
$$
\| \nabla^2 f_i(w) - \nabla^2 P(w) \| \leq \sqrt{\frac{32 L^2 \log(d/p)}{b}}.
$$
It turns out that we can consider that $\delta \sim \frac{L}{\sqrt{b}}$. Moreover, in the case when $\ell$ is a quadratic function $\delta \sim \frac{L}{b}$ \citep{hendrikx2020statistically}.}

The following theorem presents the convergence guarantees of \texttt{Shuffled-SARAH}.
\begin{theorem} \label{th1}
Suppose that Assumption \ref{ass} holds. Consider \texttt{Shuffled-SARAH} (Algorithm 1) with the choice of $\eta$ such that
\begin{equation}
\label{gamma}
\eta \leq \min \left[\frac{1}{8n L}; \frac{1}{8n^{2} \delta} \right].
\end{equation}
Then, we have
\begin{align*}
    P(w_{s+1}) - P^* + \frac{\eta (n+1)}{16} \| v_s\|^2 \leq \left( 1 - \frac{\eta \mu (n+1)}{2}\right)\left(P(w_s) - P^* + \frac{\eta (n+1)}{16} \left\|  v_{s-1} \right\|^2\right).
\end{align*}
\end{theorem}
Hence, it is easy to obtain an estimate for the number of outer iterations in Algorithm \ref{Shuffled-SARAH}. 
\begin{corollary}
Fix $\varepsilon$, and let us run \texttt{Shuffled-SARAH} with
$\eta$ from \eqref{gamma}. Then we can obtain
an $\varepsilon$-approximate solution (in terms of $P(w) - P^* \leq \varepsilon$) after
\begin{equation*}
\label{epochs}
S = \mathcal{O}\left( \left[n \cdot \frac{L}{ \mu} + n^2 \cdot \frac{\delta}{\mu} \right]\log \frac{1}{\varepsilon}\right) \quad \text{calls of terms $f_i$}.
\end{equation*}
\end{corollary}

\ab{
\paragraph{Difference between RR, SO, IG.} A crucial detail is that we achieve deterministic convergence estimates irrespective of the generation strategy $\pi_s$ employed in line \ref{randomPermutation} of Algorithm \ref{Shuffled-SARAH}. Concurrently, it's anticipated that the diverse characteristics of $\pi_s$ may result in variations in theoretical convergence. Indeed, different shuffling techniques have a theoretical impact 
on the SGD method
as discussed 
in the most recent research findings  \citep{mishchenko2020random,koloskova2023shuffle}. 
However, a significant detail of these findings is that the differences only manifest in the portion that depends on the variance bound of the stochastic gradient $\sigma^2$. But SARAH, like other methods incorporating variance reduction, nullifies the effect of $\sigma^2$. The most credible results on the effect of variance reduction and shuffling techniques do not discern a considerable difference among \textbf{RR}, \textbf{SO}, and \textbf{IG} generally \citep{malinovsky2021random,NEURIPS2021_1a3650ae}. Minor improvements are detectable in specific cases \citep{malinovsky2021random} or include only logarithmic factors \citep{NEURIPS2021_1a3650ae}. Consequently, the question of distinguishing between \textbf{RR}, \textbf{SO}, and \textbf{IG} in variance reduction algorithms remains unanswered.
}


\ab{\paragraph{Optimal choice of $b$.} In the case of $\delta \sim \frac{L}{\sqrt{b}}$, it is required to take $\frac{n}{\sqrt{b}} \sim 1$, i.e. $b = N^{2/3}$, where $N$ is the total sample size: $N = nb$.}

\section{Numerical Experiments}

\paragraph{Trajectory.}
We start with a toy experiment in $\R^{2}$ with a quadratic function. We compare the trajectories of the classical SARAH (two random and average), the average trajectory of the \texttt{RR-SARAH} (see Algorithm \ref{alg2} in Appendix \ref{rr}), and the random trajectory of \texttt{Shuffled-SARAH} with Random Reshuffling.
\begin{figure} 
        \centering
      
        \begin{minipage}{0.99\textwidth}
            \centering                
            \includegraphics[width=0.3\textwidth]{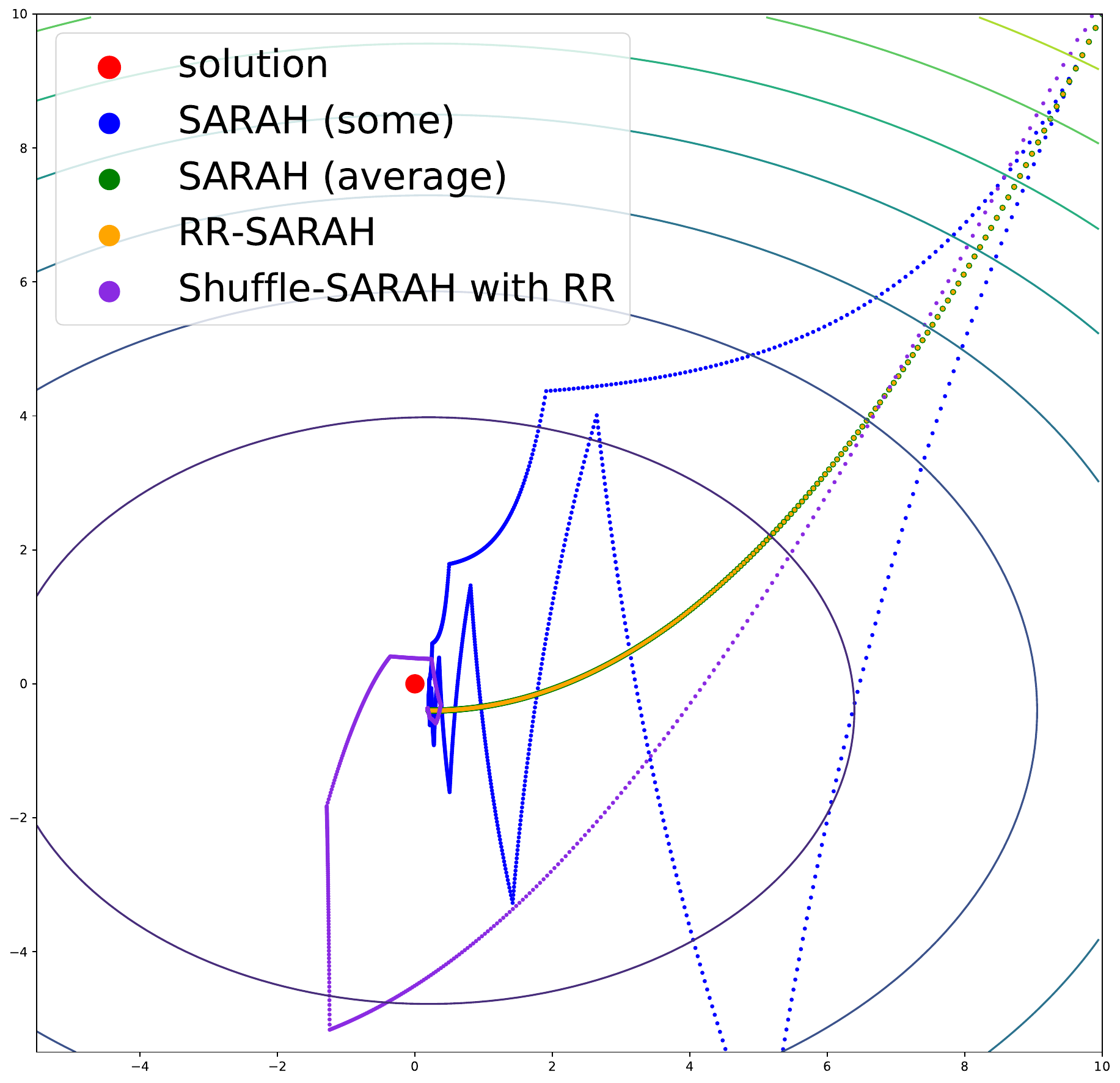}
        \end{minipage}
          \caption{Trajectories on quadratic function.}
        \label{fig:libsvm_conv}
\end{figure}
 
\paragraph{Logistic regression.}

\begin{figure}[h!]
        \centering
       
            \begin{minipage}{0.32\textwidth}
                \centering                
                \includegraphics[width=0.95\textwidth]{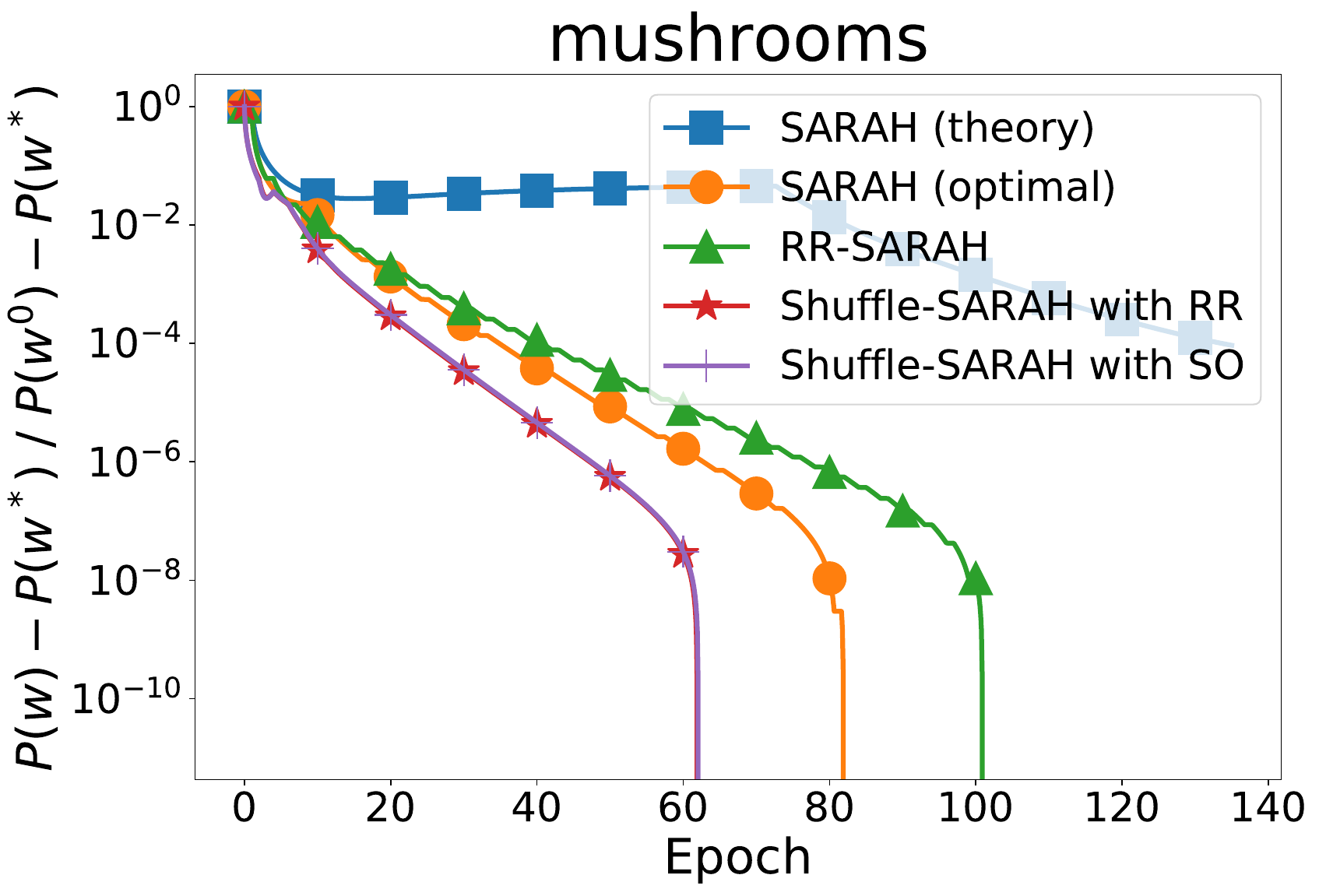}
            \end{minipage}
            \begin{minipage}{0.32\textwidth}
                \centering \includegraphics[width=0.95\textwidth]{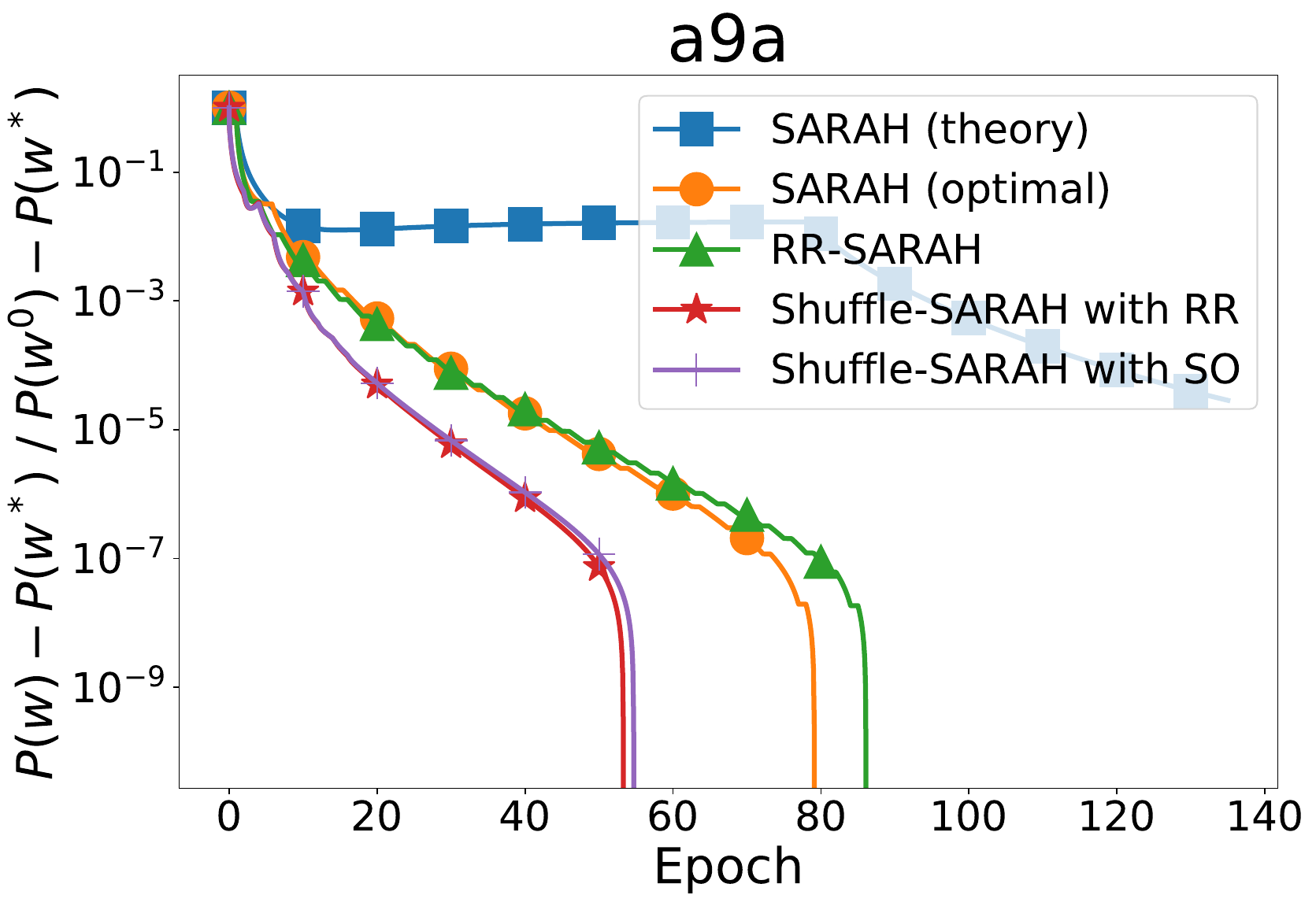}
            \end{minipage}
            \begin{minipage}{0.32\textwidth}
                \centering \includegraphics[width=0.95\textwidth]{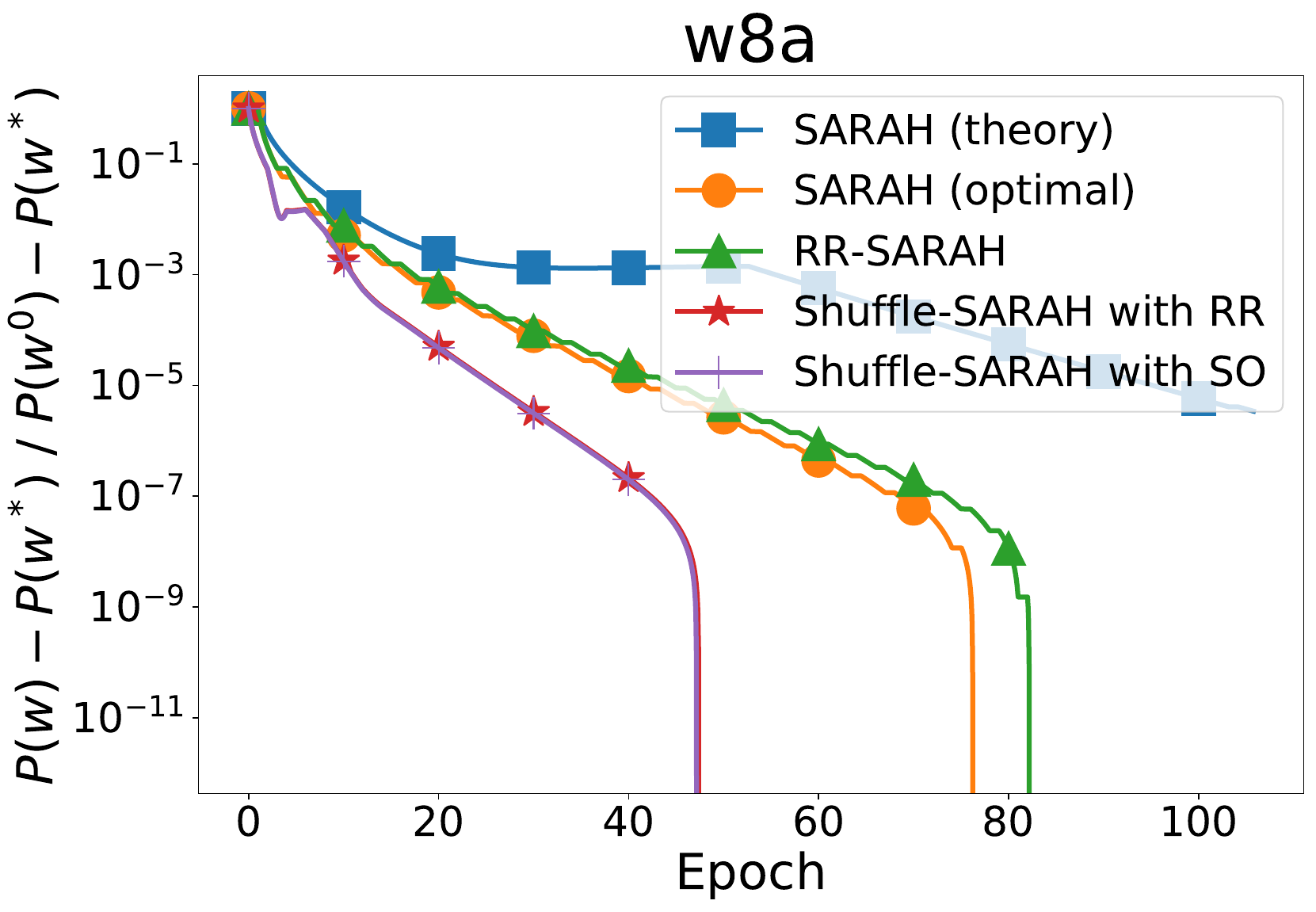}
            \end{minipage}
            \begin{minipage}{0.32\textwidth}
                \centering(a) mushrooms
            \end{minipage}
            \begin{minipage}{0.32\textwidth}
                \centering(b) a9a dataset
            \end{minipage}
            \begin{minipage}{0.32\textwidth}
                \centering(c) w8a dataset
            \end{minipage}
             \caption{Convergence of SARAH-type methods on various LiBSVM datasets. Convergence on the function.}
            \label{fig:libsvm_conv_f}
\end{figure}

Next, we consider the logistic regression problem with $\ell_2$-regularization for binary classification with
\begin{equation*}
\label{log_reg}
f_i(w) = \frac{1}{b}\sum\limits_{k=1}^b \log\left(1 + \exp\left(-y_k \cdot (X_b w)_k\right)\right) + \frac{\lambda}{2} \|w\|^2,
\end{equation*}
where $X_b \in\R^{b\times d}$ is a matrix of objects, $y_1,\ldots,y_b\in\{-1,1\}$ are labels for these objects, $b$ is the size of the local datasets and $w \in\R^d$ is a vector of weights. We optimize this problem for \texttt{mushrooms}, \texttt{a9a}, \texttt{w8a} datasets from LIBSVM library\citep{chang2011libsvm}. \ab{Since we only have the estimate on $b$ for our method (see the end of Section \ref{theory}), it is not completely fair to use it for all methods, therefore we take as $b$ the nearest power of two to $N/100$, where $N$ is the full size of the data.} More details on the dataset parameters can be found in Table \ref{tab1}.  For each dataset we take the regularization parameter $\lambda = 0.001 \cdot L$. \ab{One can note that the strong convexity constant of the problem $\mu = \lambda$.} We compare the following method settings: 1) SARAH with theoretical parameters $m = 4.5 \cdot (L/\mu)$, $\eta = 1/(2L)$ (\ab{$m$ is the number of iterations of the inner loop -- see \citep{nguyen2017sarah}}), 2) SARAH with optimal parameters (is selected by brute force -- see Table \ref{tab2}), 3) \texttt{RR-SARAH} with optimal tuned step-size, 4) \texttt{Shuffled-SARAH} (Random Reshuffling) with optimal tuned step-size, 5) \texttt{Shuffled-SARAH} (Shuffle Once) with optimal tuned step-size. All methods are run 20 times, and the convergence results are averaged. We are interested in how these methods converge in terms of the epochs number (1 epoch is a call of the full gradient of $P$). For results see Figures \ref{fig:libsvm_conv_f}, \ref{fig:libsvm_conv_x}, \ref{fig:libsvm_conv_g}. One can note that in these cases our new methods are superior to the original SARAH.

\begin{table}[t]
\centering
\caption{Summary of datasets.}
\begin{tabular}{|l|c|c|c|c|}
\hline
& $N$ & $b$ & $d$ & $L$ \\\hline
\texttt{mushrooms} &  8124 & 64 & 112 & 5,3  \\\hline
\texttt{a9a} & 32561 & 256 & 123 & 3,5 \\ \hline
\texttt{w8a} & 49749 & 256 & 300 & 28,5 \\\hline
\end{tabular}
\label{tab1}
\end{table}
\begin{table}
\centering
\caption{Optimal parameters for SARAH.}
\begin{tabular}{|l|c|c|}
\hline
& $m$ & $\eta$  \\\hline
\texttt{mushrooms} &  $0,5 \cdot (L/\mu)$ & $ 1/L$   \\\hline
\texttt{a9a} & $0,25 \cdot (L/\mu)$ & $ 1/L$ \\ \hline
\texttt{w8a} & $ L/\mu$ & $ 1/L$ \\\hline
\end{tabular}
\label{tab2}
\end{table}

\begin{figure}[h!]
        \centering
      
            \begin{minipage}{0.32\textwidth}
                \centering                
                \includegraphics[width=0.95\textwidth]{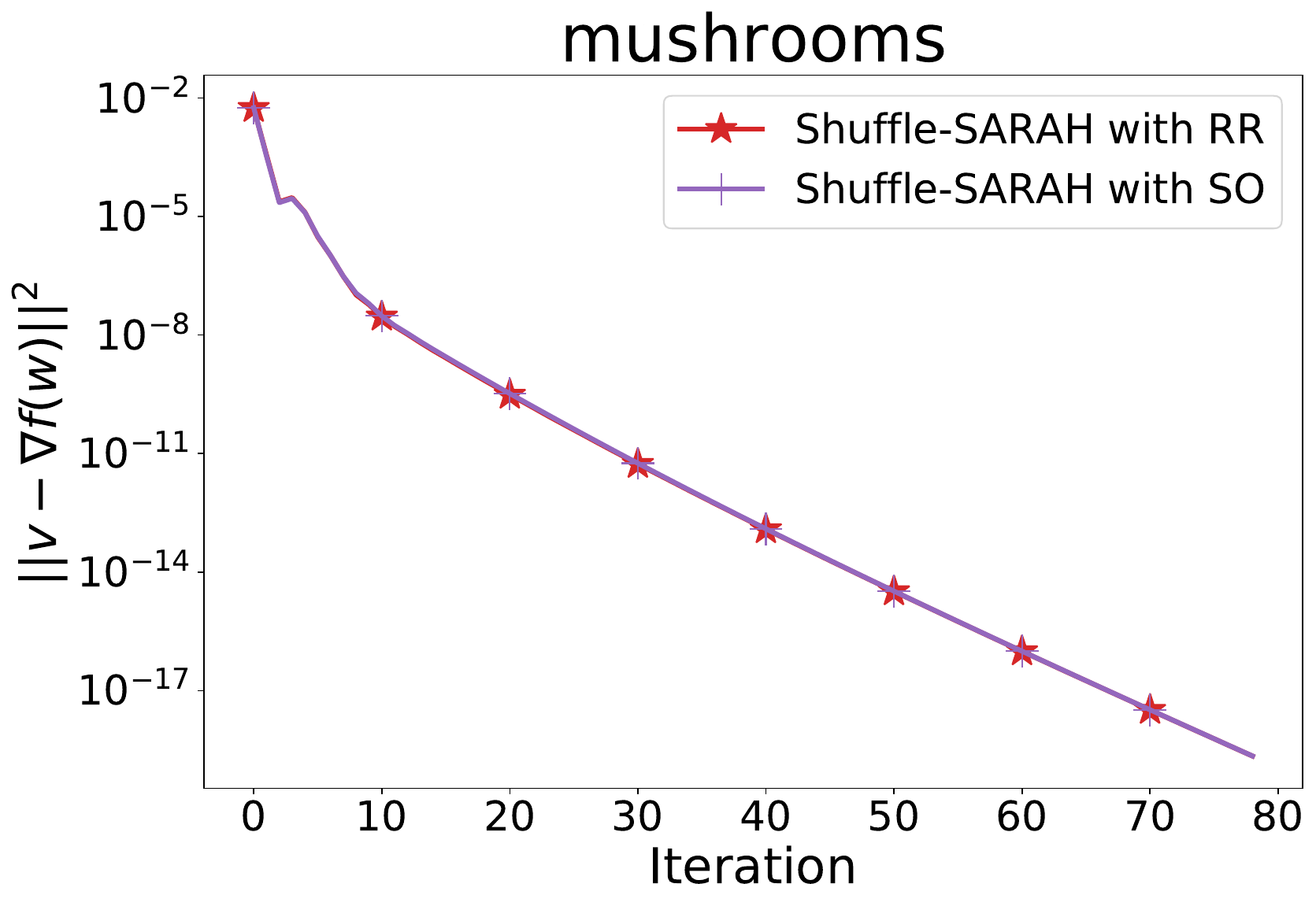}
            \end{minipage}
            \begin{minipage}{0.32\textwidth}
                \centering \includegraphics[width=0.95\textwidth]{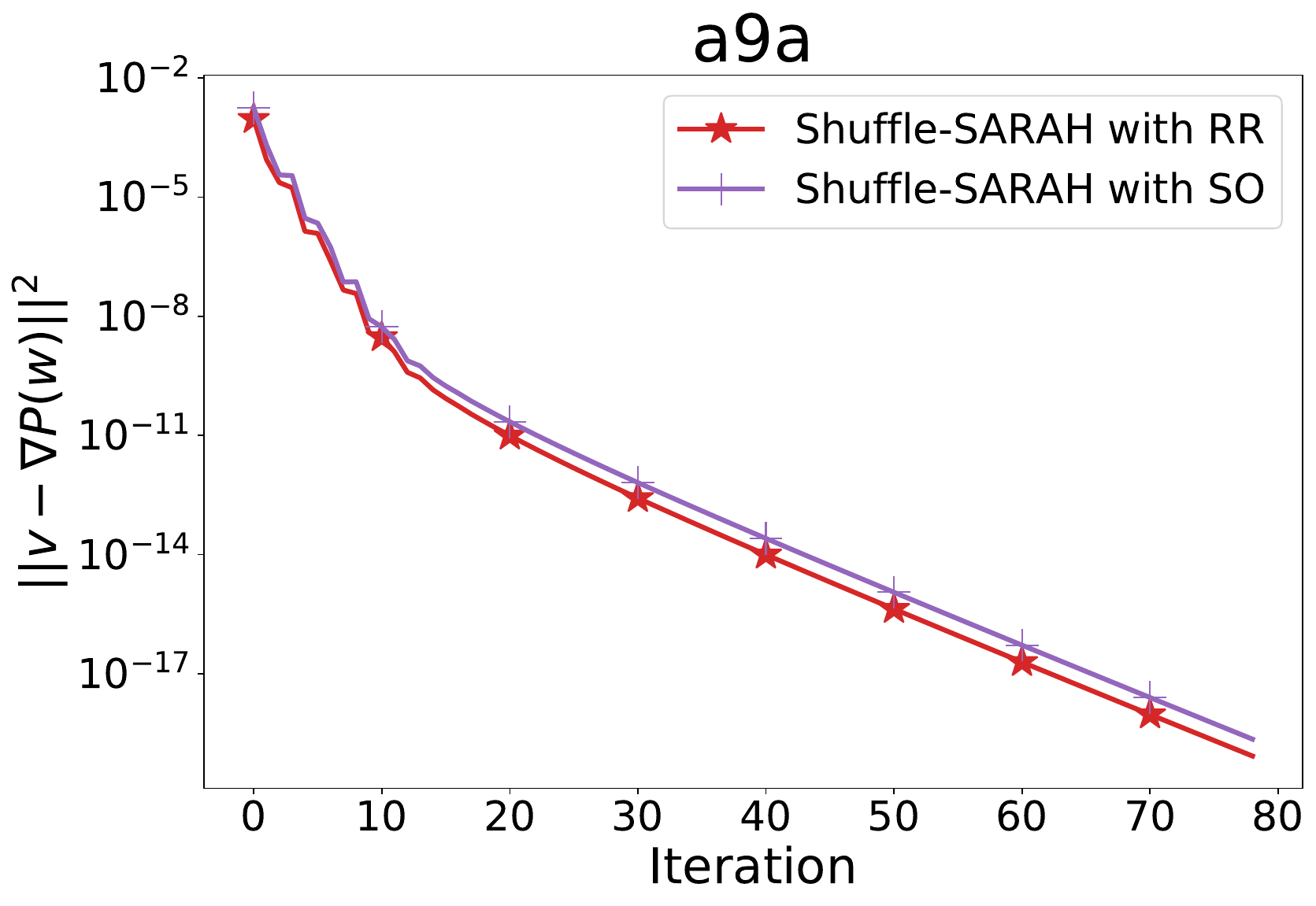}
            \end{minipage}
            \begin{minipage}{0.32\textwidth}
                \centering \includegraphics[width=0.95\textwidth]{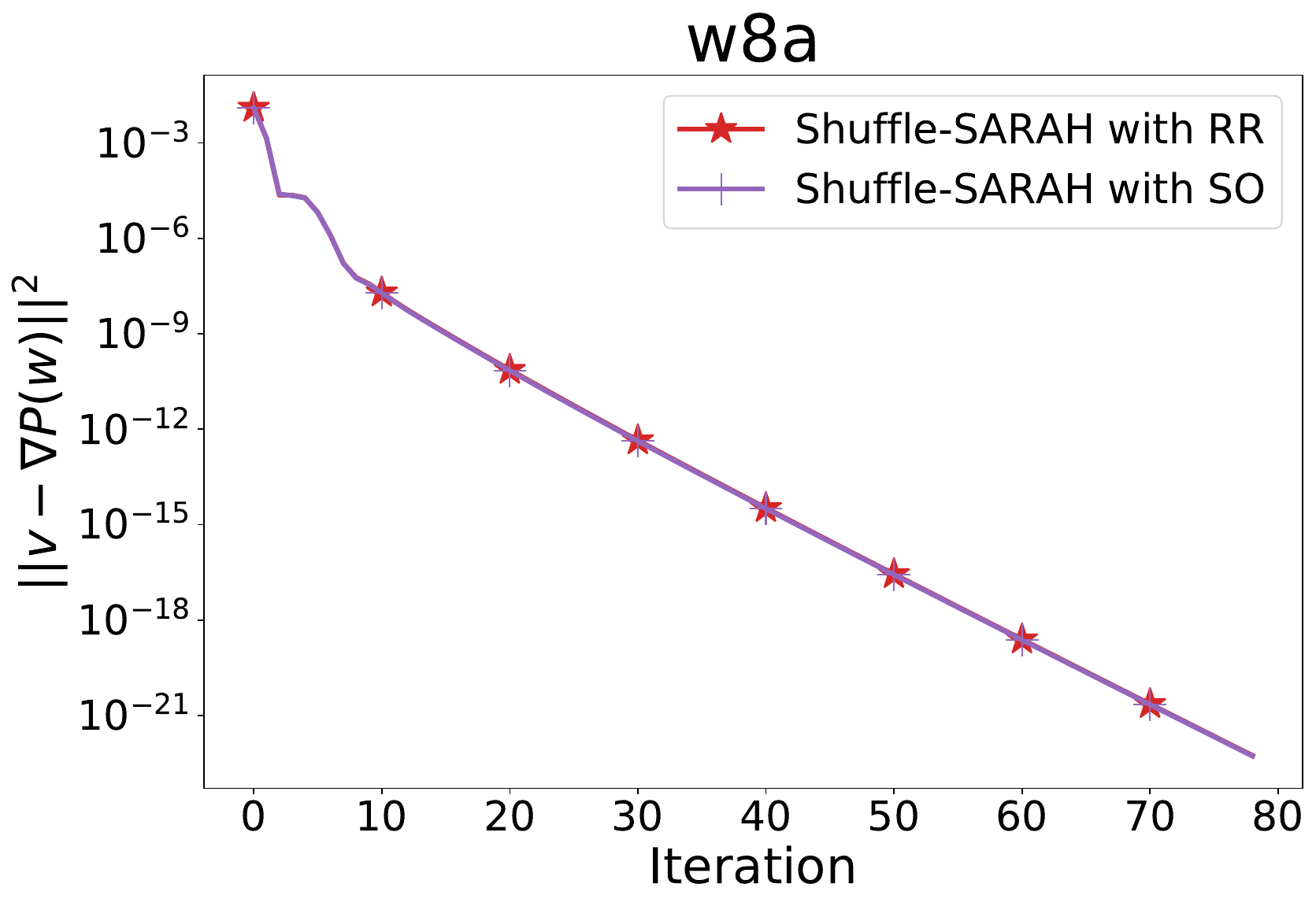}
            \end{minipage}
            \begin{minipage}{0.32\textwidth}
                \centering(a) mushrooms
            \end{minipage}
            \begin{minipage}{0.32\textwidth}
                \centering(b) a9a dataset
            \end{minipage}
            \begin{minipage}{0.32\textwidth}
                \centering(c) w8a dataset
            \end{minipage}
              \caption{$\|v_s - \nabla P(w_s)\|^2$ changes.}
            \label{fig:libsvm_grad}
\end{figure}

\paragraph{The $v_s$ is getting closer to $\nabla P(w_s)$.} 
The goal of this experiment is to show that $v$ is a good approximation of $\nabla P$ and improves with each iteration. To do this, we analyze the changes of $\|v_s - \nabla P(w_s)\|^2$ on the logistic regression problem (see the previous paragraph). See the results in Figure \ref{fig:libsvm_grad}. It can be seen that the difference is decreasing $\|v_s - \nabla P(w_s)\|^2$.

\subsection*{Acknowledgements}
The  work of A. Beznosikov  was supported by a grant for research centers in the
feld of artifcial intelligence, provided by the Analytical Center for the Government of the Russian
Federation in accordance with the subsidy agreement (agreement identifer 000000D730321P5Q0002)
and the agreement with the Moscow Institute of Physics and Technology dated November 1, 2021 No.
70-2021-00138. This work was partially conducted while A. Beznosikov, was visiting research assistants in Mohamed bin Zayed University of Artificial Intelligence (MBZUAI).

%
%

\bibliography{literature}   

\appendix

\section{Additional experimental results}

\begin{figure}[h]
        \centering
        \caption{Convergence of SARAH-type methods on various LiBSVM datasets. Convergence on the distance to the solution.}
            \begin{minipage}{0.32\textwidth}
                \centering            
                \includegraphics[width=0.95\textwidth]{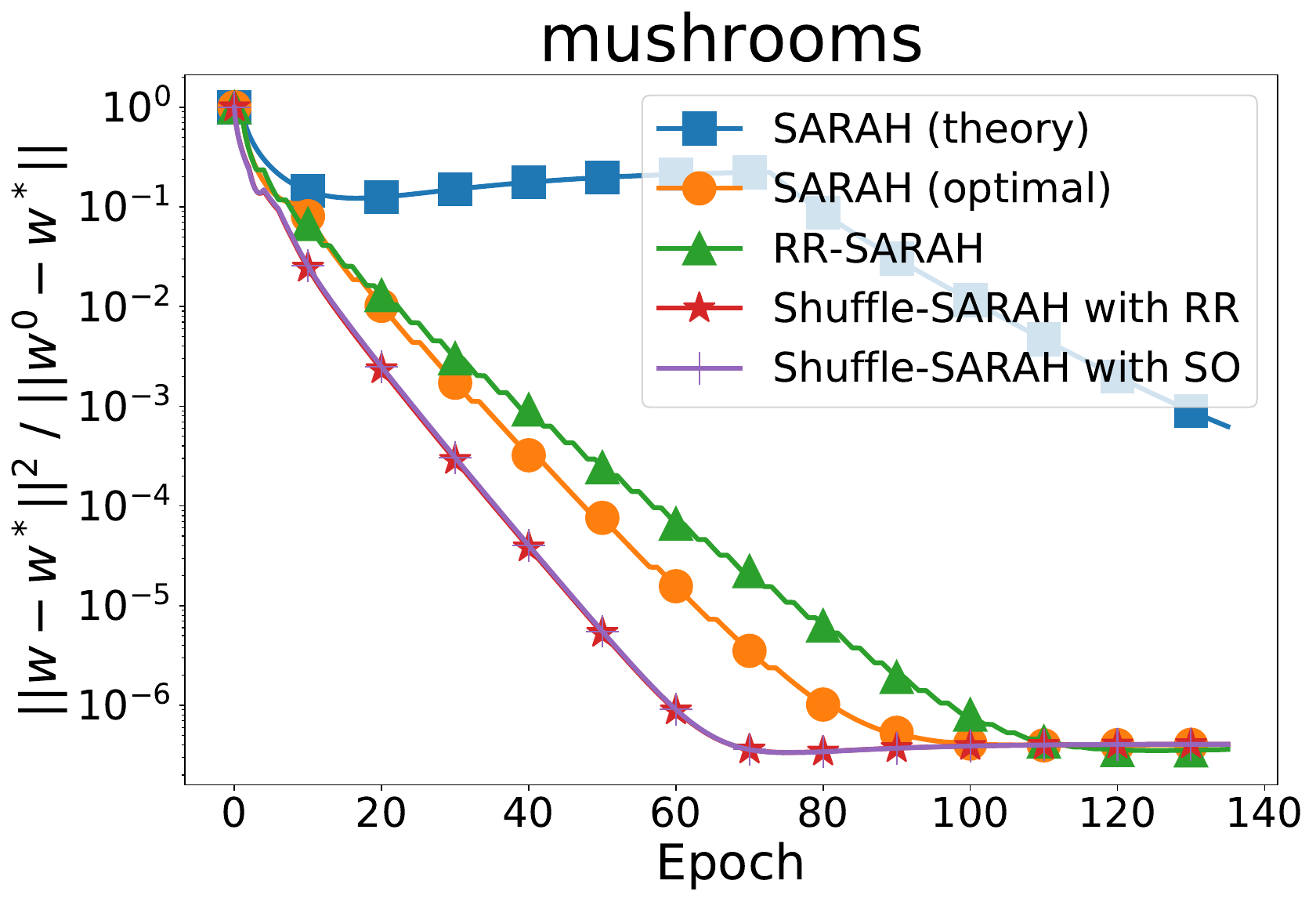}
            \end{minipage}
            \begin{minipage}{0.32\textwidth}
                \centering \includegraphics[width=0.95\textwidth]{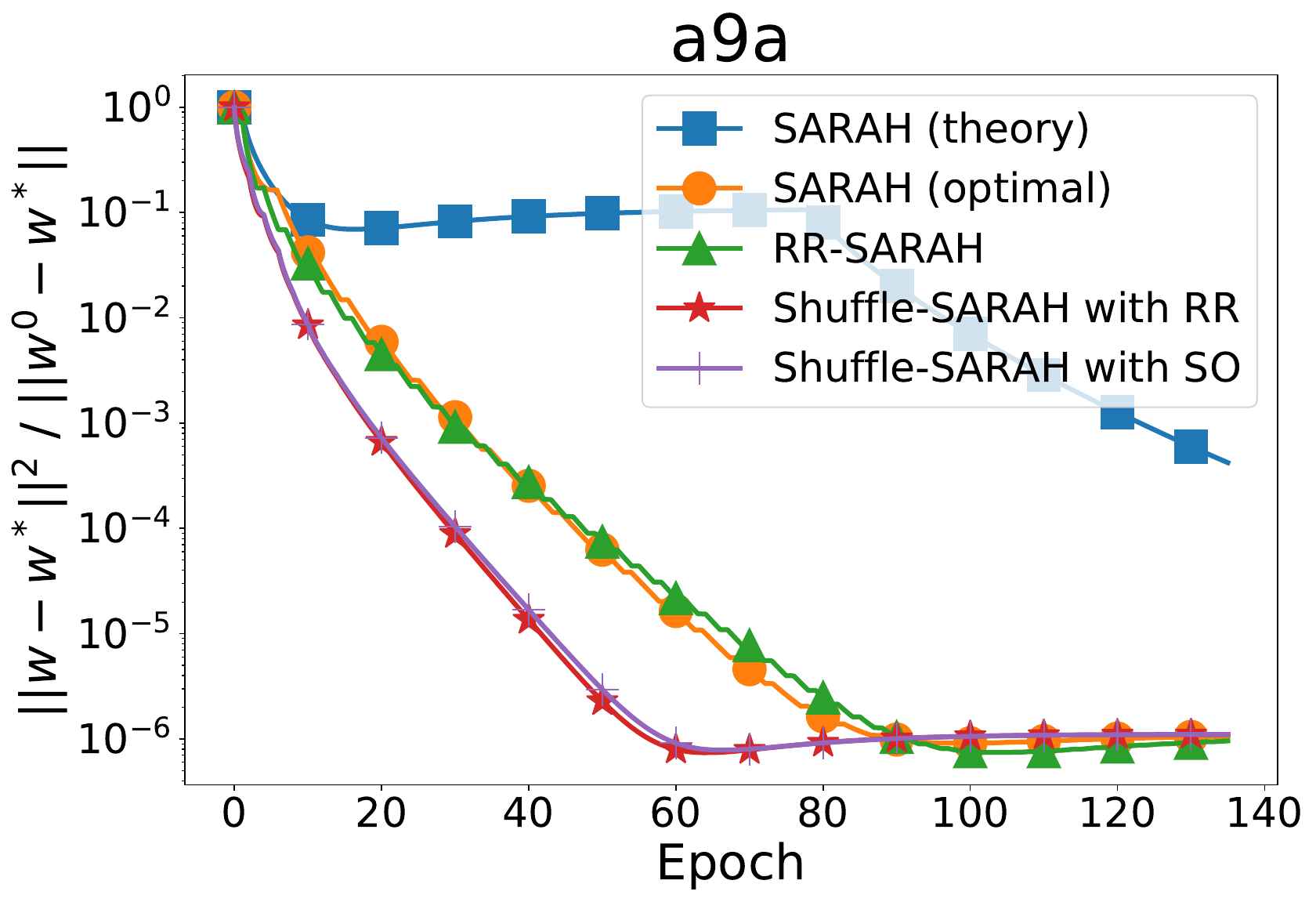}
            \end{minipage}
            \begin{minipage}{0.32\textwidth}
                \centering \includegraphics[width=0.95\textwidth]{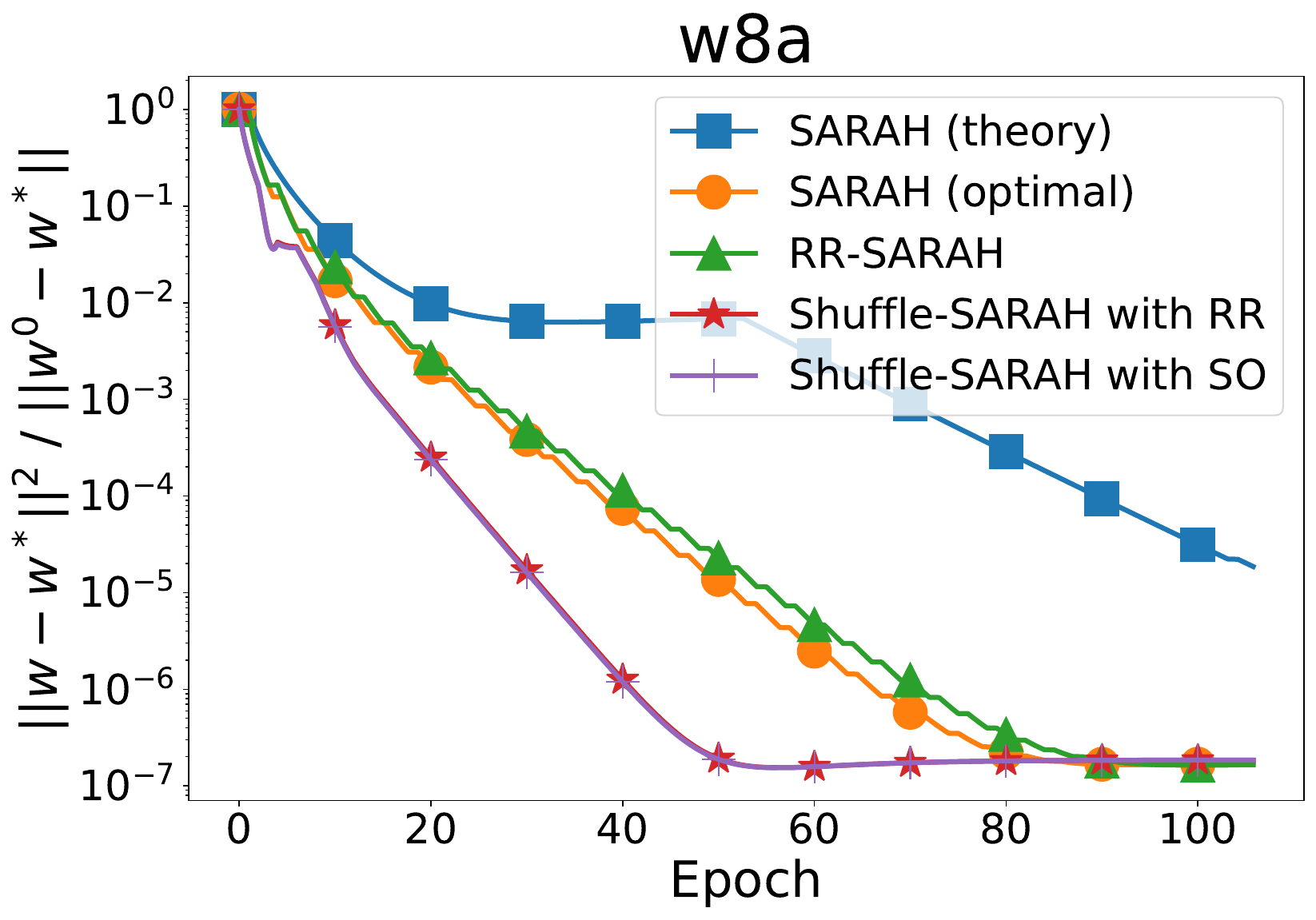}
            \end{minipage}
            \begin{minipage}{0.32\textwidth}
                \centering(a) mushrooms
            \end{minipage}
            \begin{minipage}{0.32\textwidth}
                \centering(b) a9a dataset
            \end{minipage}
            \begin{minipage}{0.32\textwidth}
                \centering(c) w8a dataset
            \end{minipage}
            \label{fig:libsvm_conv_x}
\end{figure}

\begin{figure}[h]
        \centering
        \caption{Convergence of SARAH-type methods on various LIBSVM datasets.  Convergence on the norm og the gradient.}
            \begin{minipage}{0.32\textwidth}
                \centering                
                \includegraphics[width=0.95\textwidth]{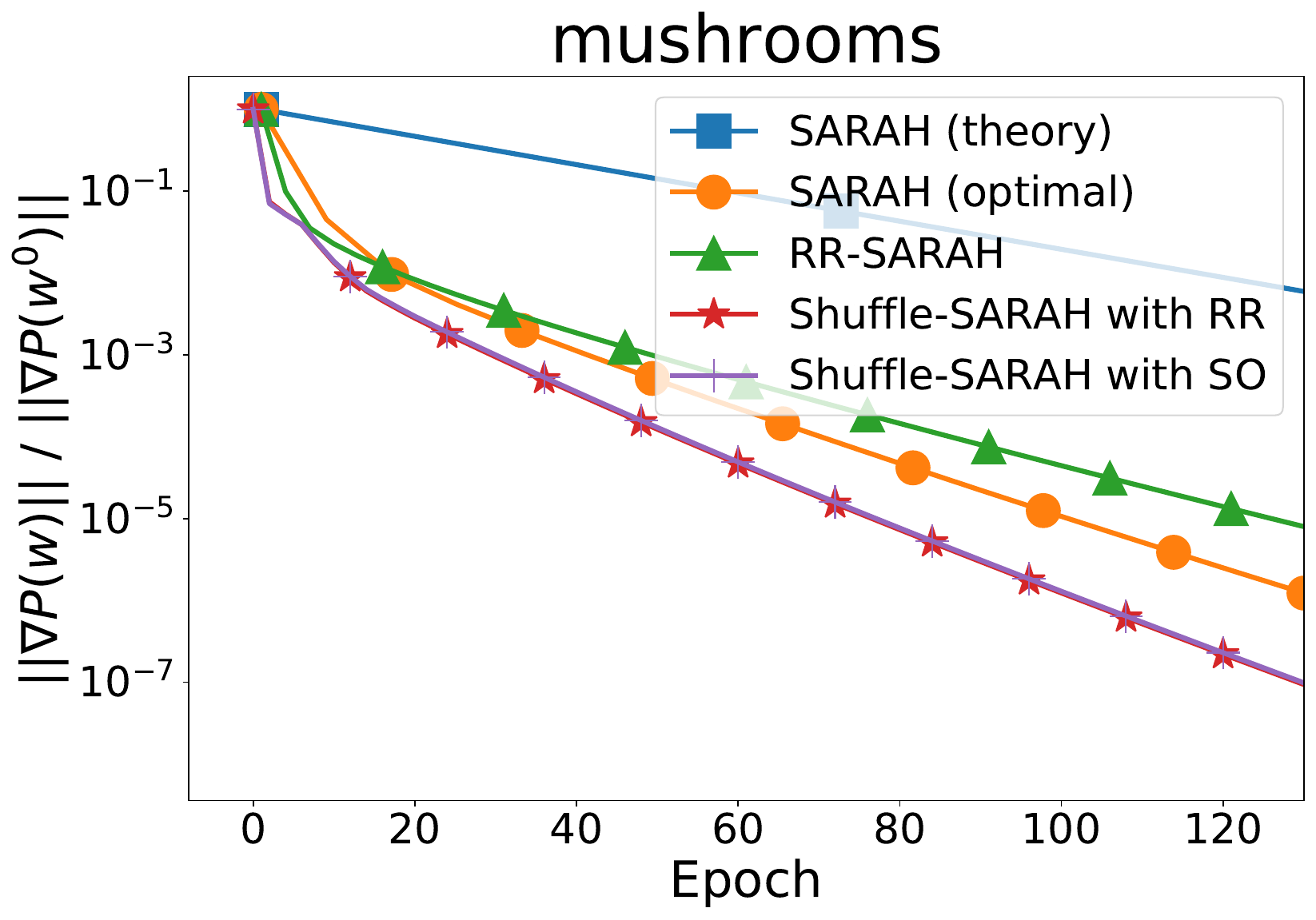}
            \end{minipage}
            \begin{minipage}{0.32\textwidth}
                \centering \includegraphics[width=0.95\textwidth]{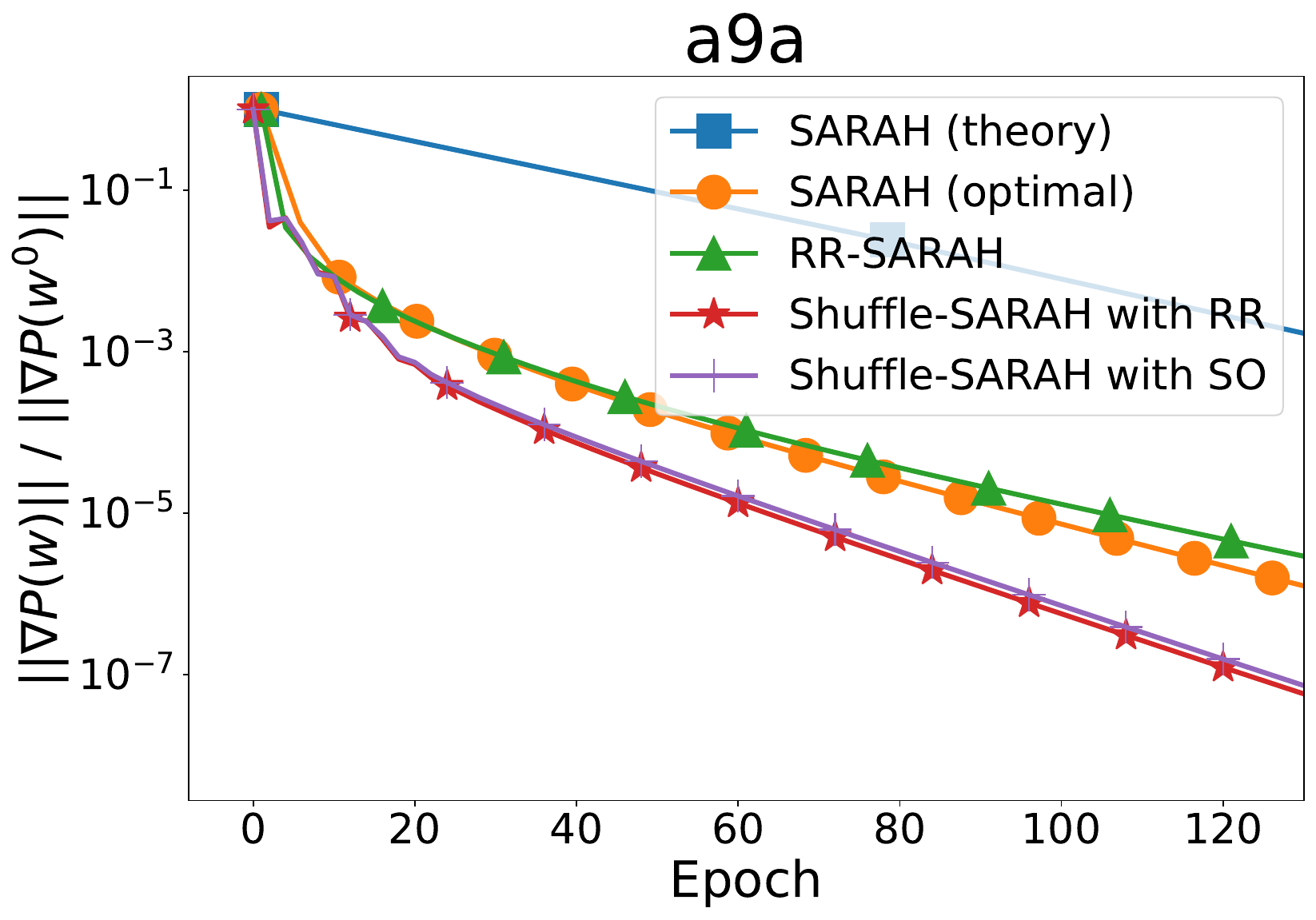}
            \end{minipage}
            \begin{minipage}{0.32\textwidth}
                \centering \includegraphics[width=0.95\textwidth]{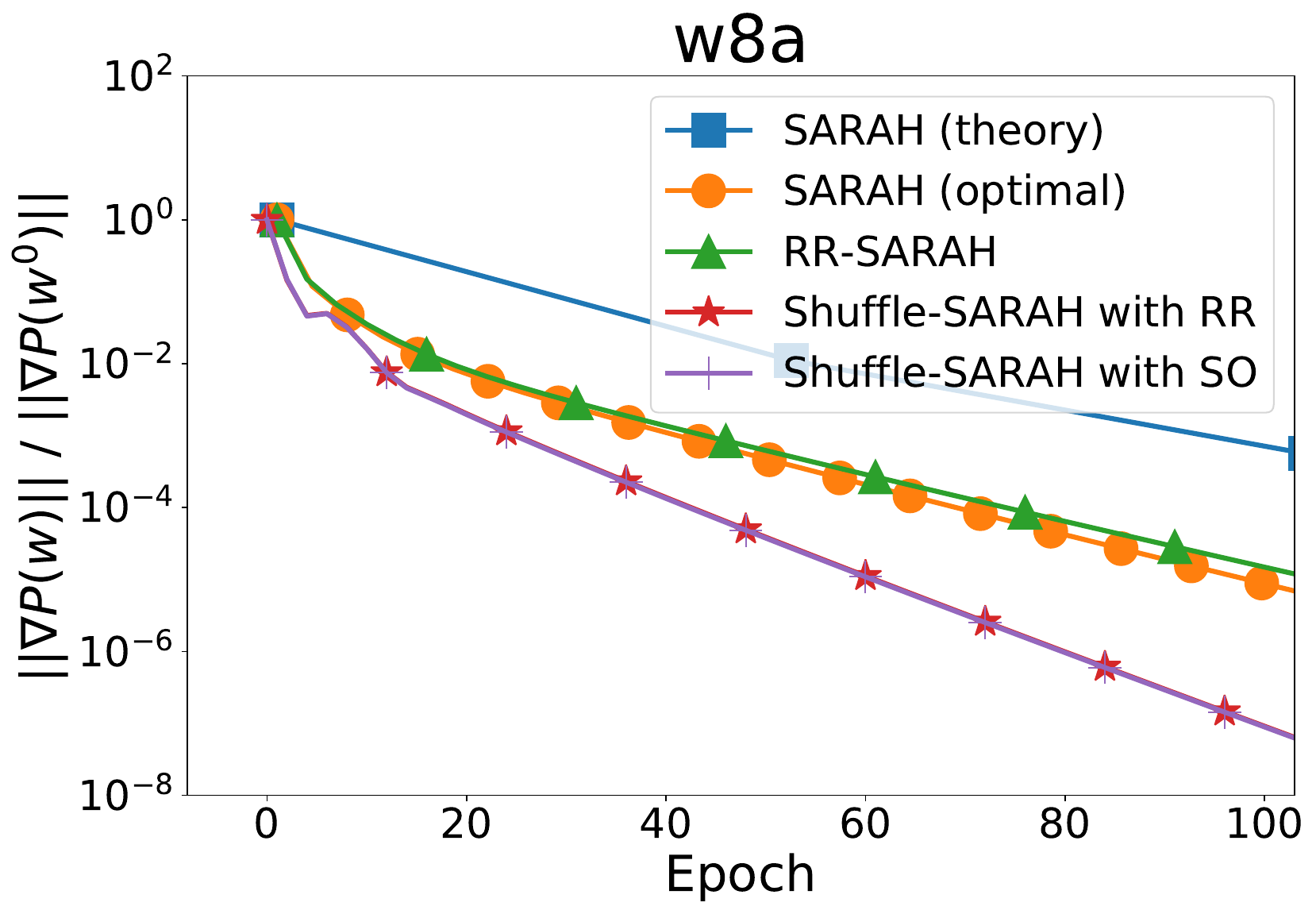}
            \end{minipage}
            \begin{minipage}{0.32\textwidth}
                \centering(a) mushrooms
            \end{minipage}
            \begin{minipage}{0.32\textwidth}
                \centering(b) a9a dataset
            \end{minipage}
            \begin{minipage}{0.32\textwidth}
                \centering(c) w8a dataset
            \end{minipage}
            \label{fig:libsvm_conv_g}
\end{figure}

\section{\texttt{RR-SARAH}} \label{rr}

This Algorithm is a modification of the original SARAH using Random Reshuffling. Unlike Algorithm \ref{Shuffled-SARAH}, this algorithm uses the full gradient $\nabla P$.

\begin{algorithm2e}[h]
\caption{\texttt{RR-SARAH}}
\label{alg2}
 \SetAlgoLined
  {\bf Input:} $0<\eta $ step-size
  
  choose $w^{-} \in \R^d$

  $w = w^{-}$

  \For{$s = 0, 1, 2, \dots$}{
      define $w_s := w$
      
      $v = \nabla P(w)$
      
      $w^- = w$
      
      $w = w - \eta v$

      sample a permutation $\pi_s = (\pi_s^1, \dots, \pi_s^n)$ of $[n]$
      
        \For{$i=1,2,\dots,n$}{

            $v = v + \nabla  f_{\pi_s^i} (w)
                             - \nabla f_{\pi_s^i} (w^{-}) $ 
                             
            $w^{-} = w$
            
            $w = w - \eta v $
     }

    } 
  
    {\bf Return: } $w$
\end{algorithm2e}

\begin{theorem} \label{rr_th}
Suppose that Assumption \ref{ass} holds. Consider \texttt{RR-SARAH} (Algorithm~\ref{alg2}) with the choice of $\eta$ such that
\begin{equation}
\label{gamma1}
\eta \leq \min \left[\frac{1}{8n L}; \frac{1}{8n^{2} \delta} \right].
\end{equation}
Then, we have
\begin{align*}
    P(w_{s+1}) - P^* &\leq \left( 1 - \frac{\eta \mu (n+1)}{2}\right)\left(P(w_s) - P^*\right).
\end{align*}
\end{theorem}
\begin{corollary}
Fix $\varepsilon$, and let us run \texttt{RR-SARAH} with
$\eta$ from \eqref{gamma1}. Then we can obtain
an $\varepsilon$-approximate solution (in terms of $P(w) - P^* \leq \varepsilon$) after
\begin{equation*}
S = \mathcal{O}\left( \left[n \cdot \frac{L}{ \mu} + n^2 \cdot \frac{\delta}{\mu} \right] \log \frac{1}{\varepsilon}\right) \quad \text{calls of terms $f_i$}.
\end{equation*}
\end{corollary}

\section{Missing proofs for Section \ref{theory} and Appendix \ref{rr}}
Before we start to prove, let us note that $\delta$-similarity from Assumption \ref{ass} gives $(\delta/2)$-smoothness of function $(f_i - P)$ for any $i \in [n]$. This implies $\delta$-smoothness of function $(f_i - f_j)$ for any $i,j \in [n]$:
\begin{align}
    \label{d-sm}
    \|\nabla f_i(w_1) &- \nabla f_j(w_1) - (\nabla f_i(w_2) - \nabla f_j(w_2))\| \nonumber\\
    &\leq \|\nabla f_i(w_1) - \nabla P(w_1) - (\nabla f_i(w_2) - \nabla P(w_2))\| \nonumber\\
    &\hspace{0.4cm}+ \|\nabla P(w_1) - \nabla f_j(w_1) - (\nabla P(w_2) - \nabla f_j(w_2))\| \nonumber\\
    &\leq 2\cdot (\delta /2)\|w_1 - w_2\|^2 = \delta\|w_1 - w_2\|^2.
\end{align}
Next, we introduce additional notation for simplicity. If we consider Algorithm \ref{Shuffled-SARAH} in iteration $s \neq 0$,  one can note that update rule is nothing more than
\begin{align}
    \label{eq:w_0} w_s &= w^0_s = w^{n+1}_{s-1}, \\
    \notag v_s &= v^0_s = \frac{1}{n} \sum\limits_{i=1}^{n}  f_{\pi^{i}_{s-1}} (w^{i}_{s-1}), \\
    \notag w^1_s &= w^0_s - \eta v^0_s, \\
    \label{eq:v_i_s} v^{i}_s &= v^{i-1}_s +  f_{\pi^{i}_{s}} (w^{i}_{s}) - f_{\pi^{i}_{s}} (w^{i-1}_{s}), \\
    \label{eq:w_i_s} w^{i+1}_s &= w^{i}_s - \eta v^{i}_s .
\end{align}
These new notations will be used further in the proofs. For Algorithm \ref{alg2}, one can do exactly the same notations with $v_s = v^0_s = \nabla P(w_s)$.

\begin{lemma} \label{lemma1}
Under Assumption \ref{ass}, for Algorithms \ref{Shuffled-SARAH} and \ref{alg2} with $\eta$ from \eqref{gamma} the following holds
\begin{align*}
    P(w_{s+1}) &\leq P(w_s) - \frac{\eta (n + 1)}{2} \| \nabla P(w_s)\|^2 + \frac{\eta (n + 1)}{2} \left\| \nabla P(w_s) - \frac{1}{n} \sum\limits_{i=0}^{n} v^i_s\right\|^2.
\end{align*}
\end{lemma}
\textbf{Proof:} Using $L$-smoothness of function $P$ (Assumption \ref{ass} (i)), we have
\begin{align*}
    P(w_{s+1}) &\leq P(w_s) + \langle \nabla P(w_s), w_{s+1} - w_s \rangle + \frac{L}{2} \|w_{s+1} - w_s \|^2 \\
    &= P(w_s) - \eta (n+1) \left\langle \nabla P(w_s), \frac{1}{n+1} \sum\limits_{i=0}^{n} v^i_s \right\rangle + \frac{\eta^2 (n+1)^2 L}{2} \left\| \frac{1}{n+1} \sum\limits_{i=0}^{n} v^i_s\right\|^2 \\
    &= P(w_s) - \frac{\eta (n+1)}{2} \left( \| \nabla P(w_s)\|^2 + \left\| \frac{1}{n+1} \sum\limits_{i=0}^{n} v^i_s\right\|^2 -  \left\| \nabla P(w_s) - \frac{1}{n+1} \sum\limits_{i=0}^{n} v^i_s\right\|^2\right) \\
    &\hspace{0.4cm}+ \frac{\eta^2 (n+1)^2 L}{2} \left\| \frac{1}{n+1} \sum\limits_{i=0}^{n} v^i_s\right\|^2 \\
    &= P(w_s) - \frac{\eta (n+1)}{2} \| \nabla P(w_s)\|^2 - \frac{\eta (n+1)}{2} (1 - \eta (n+1) L) \left\| \frac{1}{n+1} \sum\limits_{i=0}^{n} v^i_s\right\|^2 
    \\
    &\hspace{0.4cm}+ \frac{\eta (n+1)}{2} \left\| \nabla P(w_s) - \frac{1}{n+1} \sum\limits_{i=0}^{n} v^i_s\right\|^2.
\end{align*}
With $\eta \leq \frac{1}{8nL} \leq  \frac{1}{(n+1)L}$, we get
\begin{align*}
    P(w_{s+1}) &\leq P(w_s) - \frac{\eta (n+1)}{2} \| \nabla P(w_s)\|^2 + \frac{\eta (n+1)}{2} \left\| \nabla P(w_s) - \frac{1}{n+1} \sum\limits_{i=0}^{n} v^i_s\right\|^2.
\end{align*}
Which completes the proof. 
\EndProof

\begin{lemma} \label{lemma2}
Under Assumption \ref{ass}, for Algorithms \ref{Shuffled-SARAH} and \ref{alg2} the following holds
\begin{align*}
     \left\| \nabla P(w_s) - \frac{1}{n+1} \sum\limits_{i=0}^{n} v^i_s\right\|^2 
     \leq 2\| \nabla P(w_s) - v_s \|^2 + \left( \frac{4L^2}{n+1} + 4 \delta^2 n\right) \sum\limits_{i=1}^n \|w^{i}_s - w_s\|^2.  
\end{align*}
\end{lemma}
\ab{
\textbf{Proof:} To begin with, we prove that for any $k = n, \ldots, 0$, it holds
\begin{align}
    \label{eq:lem2:ind1}
    \sum\limits_{i=k}^{n} v^i_s =& \sum\limits_{i=k+1}^n \left((n+1-i)\cdot\left[ \nabla f_{\pi^{i}_s} (w^{i}_s) -\nabla f_{\pi^{i}_s} (w^{i-1}_s) \right] \right) + (n - k +1)v^k_s.
\end{align}
One can prove it by mathematical induction. For $k = n$, we have $\sum\limits_{i=n}^{n} v^i_s = v^n_s$. Suppose \eqref{eq:lem2:ind1} holds true for $k$, let us prove for $k-1$:
\begin{align*}
    \sum\limits_{i=k-1}^{n} v^i_s &= v^{k-1}_s + \sum\limits_{i=k}^{n} v^i_s
    \\
    &= v^{k-1}_s + \sum\limits_{i=k+1}^n \left((n+1-i)\cdot\left[ \nabla f_{\pi^{i}_s} (w^{i}_s) -\nabla f_{\pi^{i}_s} (w^{i-1}_s) \right] \right) + (n - k +1)v^k_s
    \\
    &= v^{k-1}_s + \sum\limits_{i=k+1}^n \left((n+1-i)\cdot\left[ \nabla f_{\pi^{i}_s} (w^{i}_s) -\nabla f_{\pi^{i}_s} (w^{i-1}_s) \right] \right) 
    \\
    &\hspace{0.4cm}
    + (n - k +1) \left[v^{k-1}_s +  f_{\pi^{k}_{s}} (w^{k}_{s}) - f_{\pi^{k}_{s}} (w^{k-1}_{s})\right]
    \\
    &{=} \sum\limits_{i= k}^n \left((n+1-i)\cdot\left[ \nabla f_{\pi^{i}_s} (w^{i}_s) -\nabla f_{\pi^{i}_s} (w^{i-1}_s) \right] \right) 
    + (n - k) v^{k-1}_s .
\end{align*}
Here we additionally used \eqref{eq:v_i_s}. This completes the proof of \eqref{eq:lem2:ind1}. In particular, \eqref{eq:lem2:ind1} with $k = 0$ gives
\begin{align*}
     \Bigg\| \nabla P(w_s) &- \frac{1}{n+1} \sum\limits_{i=0}^{n} v^i_s\Bigg\|^2 \\
     &= \frac{1}{(n+1)^2}\left\| (n+1)\nabla P(w_s) - \sum\limits_{i=0}^{n} v^i_s\right\|^2 \\
     &= \frac{1}{(n+1)^2}\Bigg\| (n+1)\nabla P(w_s) \\
     &\hspace{0.4cm}- \sum\limits_{i=1}^n \left((n+1-i)\cdot\left[ \nabla f_{\pi^{i}_s} (w^{i}_s) -\nabla f_{\pi^{i}_s} (w^{i-1}_s) \right] \right) + (n +1)v^0_s\Bigg\|^2.
\end{align*}
Using $\| a + b\|^2 \leq 2\| a\|^2 + 2\|b\|^2$, we get
\begin{align}
    \label{eq:lem2:temp1}
     \Bigg\| \nabla P(w_s) &- \frac{1}{n+1} \sum\limits_{i=0}^{n} v^i_s\Bigg\|^2
     \notag\\
     &\leq 
     2 \| \nabla P(w_s) - v^0_s\|^2
     \notag\\
     &\hspace{0.4cm}+ 
     \frac{2}{(n+1)^2}\left\| \sum\limits_{i=1}^n \left((n+1-i)\cdot\left[ \nabla f_{\pi^{i}_s} (w^{i}_s) -\nabla f_{\pi^{i}_s} (w^{i-1}_s) \right] \right)\right\|^2.
\end{align}
Again using mathematical induction, we prove for $k = n, \ldots, 1$ the following estimate:
\begin{align}
    \label{eq:lem2:ind2}
     \Bigg\| &\sum\limits_{i=1}^n \left((n+1-i)\cdot\left[ \nabla f_{\pi^{i}_s} (w^{i}_s) -\nabla f_{\pi^{i}_s} (w^{i-1}_s) \right] \right) \Bigg\|^2
     \notag\\
     &\hspace{1cm}\leq  2 L^2 (n+1) \sum\limits_{i=k+1}^n \|w^{i}_s  - w_s\big\|^2  + 2 \delta^2 (n+1) \sum\limits_{i=k+1}^n (n + 1 - i)^2 \|w_s - w^{i-1}_s \|^2
     \notag\\
     &\hspace{1.4cm}+\frac{n+1}{k+1} \Bigg\|(n-k)\nabla f_{\pi^{k}_s} (w_s) + \nabla f_{\pi^{k}_s} (w^{k}_s) - (n-k+1)\nabla f_{\pi^{k}_s} (w^{k-1}_s)\notag\\
     &\hspace{1.4cm}+ \sum\limits_{i=1}^{k-1} \left((n+1-i)\cdot\left[ \nabla f_{\pi^{i}_s} (w^{i}_s) -\nabla f_{\pi^{i}_s} (w^{i-1}_s) \right] \right) \Bigg\|^2.
\end{align}
For $k = n$, the statement holds automatically. Suppose \eqref{eq:lem2:ind2} holds true for $k$, let us prove for $k-1$:
\begin{align*}
     \Bigg\| &\sum\limits_{i=1}^n \left((n+1-i)\cdot\left[ \nabla f_{\pi^{i}_s} (w^{i}_s) -\nabla f_{\pi^{i}_s} (w^{i-1}_s) \right] \right)\Bigg\|^2
     \notag\\
     &\hspace{1cm}\leq  2 L^2 (n+1) \sum\limits_{i=k+1}^n \|w^{i}_s  - w_s\big\|^2  + 2 \delta^2 (n+1) \sum\limits_{i=k+1}^n (n - i + 1)^2 \|w_s - w^{i-1}_s \|^2
     \notag\\
     &\hspace{1.4cm}+\frac{n+1}{k+1} \Bigg\|(n-k)\nabla f_{\pi^{k}_s} (w_s) + \nabla f_{\pi^{k}_s} (w^{k}_s) - (n-k+1)\nabla f_{\pi^{k}_s} (w^{k-1}_s)\notag\\
     &\hspace{1.4cm}+\sum\limits_{i=1}^{k-1} \left((n+1-i)\cdot\left[ \nabla f_{\pi^{i}_s} (w^{i}_s) -\nabla f_{\pi^{i}_s} (w^{i-1}_s) \right] \right) \Bigg\|^2
     \\
     &\hspace{1cm}=2 L^2 (n+1) \sum\limits_{i=k+1}^n \|w^{i}_s  - w_s\big\|^2  + 2 \delta^2 (n+1) \sum\limits_{i=k+1}^n (n - i + 1)^2 \|w_s - w^{i-1}_s \|^2
     \notag\\
     &\hspace{1.4cm}+\frac{n+1}{k+1} \Bigg\|\nabla f_{\pi^{k}_s} (w^{k}_s) - f_{\pi^{k}_s} (w_s) + (n-k+1)\nabla f_{\pi^{k}_s} (w_s) - (n-k+1)\nabla f_{\pi^{k}_s} (w^{k-1}_s)
     \\
     &\hspace{1.4cm}+ (n-k+2)\cdot\left[ \nabla f_{\pi^{k-1}_s} (w^{k-1}_s) -\nabla f_{\pi^{k-1}_s} (w^{k-2}_s) \right]
     \\
     &\hspace{1.4cm}+ \sum\limits_{i=1}^{k-2} \left((n+1-i)\cdot\left[ \nabla f_{\pi^{i}_s} (w^{i}_s) -\nabla f_{\pi^{i}_s} (w^{i-1}_s) \right] \right) \Bigg\|^2
     \\
     &\hspace{1cm}=2 L^2 (n+1) \sum\limits_{i=k+1}^n \|w^{i}_s  - w_s\big\|^2  + 2 \delta^2 (n+1) \sum\limits_{i=k+1}^n (n - i + 1)^2 \|w_s - w^{i-1}_s \|^2
     \notag\\
     &\hspace{1.4cm}+\frac{n+1}{k+1} \Bigg\|\nabla f_{\pi^{k}_s} (w^{k}_s) - f_{\pi^{k}_s} (w_s) 
     \\
     &\hspace{1.4cm}+ (n-k+1) \cdot \left[\nabla f_{\pi^{k}_s} (w_s) - \nabla f_{\pi^{k-1}_s} (w_s)  - \nabla f_{\pi^{k}_s} (w^{k-1}_s) + \nabla f_{\pi^{k-1}_s} (w^{k-1}_s) \right]
     \\
     &\hspace{1.4cm}
     + (n-k+1)\nabla f_{\pi^{k-1}_s} (w_s)  + \nabla f_{\pi^{k-1}_s} (w^{k-1}_s) -(n-k+2) \nabla f_{\pi^{k-1}_s} (w^{k-2}_s)
     \\
     &\hspace{1.4cm}+ \sum\limits_{i=1}^{k-2} \left((n+1-i)\cdot\left[ \nabla f_{\pi^{i}_s} (w^{i}_s) -\nabla f_{\pi^{i}_s} (w^{i-1}_s) \right] \right) \Bigg\|^2.
\end{align*}
Using $\| a + b\|^2 \leq (1 + c)\| a\|^2 + (1+ 1/c)\|b\|^2$ with $c = k$, we have
\begin{align*}
     \Bigg\| &\sum\limits_{i=1}^n \left((n+1-i)\cdot\left[ \nabla f_{\pi^{i}_s} (w^{i}_s) -\nabla f_{\pi^{i}_s} (w^{i-1}_s) \right] \right)\Bigg\|^2
     \notag\\
     &\hspace{1cm}\leq 2 L^2 (n+1) \sum\limits_{i=k+1}^n \|w^{i}_s  - w_s\big\|^2  + 2 \delta^2 (n+1) \sum\limits_{i=k+1}^n (n - i + 1)^2 \|w_s - w^{i-1}_s \|^2
     \notag\\
     &\hspace{1.4cm}+(n+1) \Big\|\nabla f_{\pi^{k}_s} (w^{k}_s) - f_{\pi^{k}_s} (w_s) 
     \\
     &\hspace{1.4cm}+ (n-k+1) \cdot \left[\nabla f_{\pi^{k}_s} (w_s) - \nabla f_{\pi^{k-1}_s} (w_s)  - \nabla f_{\pi^{k}_s} (w^{k-1}_s) + \nabla f_{\pi^{k-1}_s} (w^{k-1}_s) \right]\Big\|^2
     \\
     &\hspace{1.4cm}
     + \frac{n+1}{k} \Bigg\| (n-k+1)\nabla f_{\pi^{k-1}_s} (w_s)  + \nabla f_{\pi^{k-1}_s} (w^{k-1}_s) -(n-k+2) \nabla f_{\pi^{k-1}_s} (w^{k-2}_s)
     \\
     &\hspace{1.4cm}+ \sum\limits_{i=1}^{k-2} \left((n+1-i)\cdot\left[ \nabla f_{\pi^{i}_s} (w^{i}_s) -\nabla f_{\pi^{i}_s} (w^{i-1}_s) \right] \right) \Bigg\|^2.
\end{align*}
With $\| a + b\|^2 \leq 2\| a\|^2 + 2\|b\|^2$ Assumption \ref{ass} ($\delta$-similarity \eqref{d-sm} and $L$-smoothness), one can obtain
\begin{align*}
     \Bigg\| &\sum\limits_{i=1}^n \left((n+1-i)\cdot\left[ \nabla f_{\pi^{i}_s} (w^{i}_s) -\nabla f_{\pi^{i}_s} (w^{i-1}_s) \right] \right)\Bigg\|^2
     \notag\\
     &\hspace{1cm}\leq 2 L^2 (n+1) \sum\limits_{i=k+1}^n \|w^{i}_s  - w_s\big\|^2  + 2 \delta^2 (n+1) \sum\limits_{i=k+1}^n (n - i + 1)^2 \|w_s - w^{i-1}_s \|^2
     \notag\\
     &\hspace{1.4cm}+2(n+1) \|\nabla f_{\pi^{k}_s} (w^{k}_s) - f_{\pi^{k}_s} (w_s)  \|^2
     \\
     &\hspace{1.4cm}+ 2 (n+1) (n-k+1)^2 \| \nabla f_{\pi^{k}_s} (w_s) - \nabla f_{\pi^{k-1}_s} (w_s)  - \nabla f_{\pi^{k}_s} (w^{k-1}_s) + \nabla f_{\pi^{k-1}_s} (w^{k-1}_s) \|^2
     \\
     &\hspace{1.4cm}
     + \frac{n+1}{k} \Bigg\| (n-k+1)\nabla f_{\pi^{k-1}_s} (w_s)  + \nabla f_{\pi^{k-1}_s} (w^{k-1}_s) -(n-k+2) \nabla f_{\pi^{k-1}_s} (w^{k-2}_s)
     \\
     &\hspace{1.4cm}+ \sum\limits_{i=1}^{k-2} \left((n+1-i)\cdot\left[ \nabla f_{\pi^{i}_s} (w^{i}_s) -\nabla f_{\pi^{i}_s} (w^{i-1}_s) \right] \right) \Bigg\|^2
     \notag\\
     &\hspace{1cm}\leq 2 L^2 (n+1) \sum\limits_{i=k}^n \|w^{i}_s  - w_s\big\|^2  + 2 \delta^2 (n+1) \sum\limits_{i=k}^n (n - i + 1)^2 \|w_s - w^{i-1}_s \|^2
     \\
     &\hspace{1.4cm}
     + \frac{n+1}{k} \Bigg\| (n-k+1)\nabla f_{\pi^{k-1}_s} (w_s)  + \nabla f_{\pi^{k-1}_s} (w^{k-1}_s) -(n-k+2) \nabla f_{\pi^{k-1}_s} (w^{k-2}_s)
     \\
     &\hspace{1.4cm}+ \sum\limits_{i=1}^{k-2} \left((n+1-i)\cdot\left[ \nabla f_{\pi^{i}_s} (w^{i}_s) -\nabla f_{\pi^{i}_s} (w^{i-1}_s) \right] \right) \Bigg\|^2.
\end{align*}
This completes the proof of \eqref{eq:lem2:ind2}. In particular, \eqref{eq:lem2:ind2} with $k = 1$ gives
\begin{align*}
     \Bigg\| &\sum\limits_{i=1}^n \left((n+1-i)\cdot\left[ \nabla f_{\pi^{i}_s} (w^{i}_s) -\nabla f_{\pi^{i}_s} (w^{i-1}_s) \right] \right) \Bigg\|^2
     \notag\\
     &\hspace{1cm}\leq  2 L^2 (n+1) \sum\limits_{i=2}^n \|w^{i}_s  - w_s\big\|^2  + 2 \delta^2 (n+1) \sum\limits_{i=2}^n (n - i + 1)^2 \|w_s - w^{i-1}_s \|^2
     \notag\\
     &\hspace{1.4cm}+\frac{n+1}{2} \|(n-1)\nabla f_{\pi^{1}_s} (w_s) + \nabla f_{\pi^{1}_s} (w^{1}_s) - n\nabla f_{\pi^{1}_s} (w^{0}_s)\|^2.
\end{align*}
With \eqref{eq:w_0} and $L$-smoothness of function $f_{\pi^{1}_s}$ (Assumption \ref{ass} (i)), we have
\begin{align}
\label{eq:lem2:temp2}
     \Bigg\| &\sum\limits_{i=1}^n \left((n+1-i)\cdot\left[ \nabla f_{\pi^{i}_s} (w^{i}_s) -\nabla f_{\pi^{i}_s} (w^{i-1}_s) \right] \right)\Bigg\|^2
     \notag\\
     &\hspace{1cm}\leq  2 L^2 (n+1) \sum\limits_{i=2}^n \|w^{i}_s  - w_s\big\|^2  + 2 \delta^2 (n+1) \sum\limits_{i=2}^n (n - i + 1)^2 \|w_s - w^{i-1}_s \|^2
     \notag\\
     &\hspace{1.4cm}+\frac{n+1}{2} \|\nabla f_{\pi^{1}_s} (w^{1}_s) - \nabla f_{\pi^{1}_s} (w_s)\|^2
     \notag\\
     &\hspace{1cm}\leq  2 L^2 (n+1) \sum\limits_{i=1}^n \|w^{i}_s  - w_s\big\|^2  + 2 \delta^2 (n+1) \sum\limits_{i=2}^n (n - i + 1)^2 \|w_s - w^{i-1}_s \|^2
     \notag\\
     &\hspace{1cm}\leq  \left( 2 L^2 (n+1) + 2 \delta^2 (n+1) n^2 \right)\sum\limits_{i=1}^n \|w^{i}_s  - w_s\big\|^2.
\end{align}
Substituting of \eqref{eq:lem2:temp2} to \eqref{eq:lem2:temp1} completes proof.
\EndProof
}

\begin{lemma} \label{lemma3}
Under Assumption \ref{ass}, for Algorithms \ref{Shuffled-SARAH} and \ref{alg2} with $\eta$ from \eqref{gamma} the following holds for $i \in [n]$
\begin{align*}
    \|v^i_s\|^2 \leq \|v^{i-1}_s\|^2.  
\end{align*}
\end{lemma}
\textbf{Proof:} With \eqref{eq:v_i_s} and \eqref{eq:w_i_s}, we have
\begin{align*}
    \| v^{i}_s \|^2 &= \| v^{i-1}_s +  f_{\pi^{i}_{s}} (w^{i}_{s}) - f_{\pi^{i}_{s}} (w^{i-1}_{s}) \|^2
    \\
    &= \| v^{i-1}_s \|^2 +  \|f_{\pi^{i}_{s}} (w^{i}_{s}) - f_{\pi^{i}_{s}} (w^{i-1}_{s}) \|^2
    + 2\langle v^{i-1}_s, f_{\pi^{i}_{s}} (w^{i}_{s}) - f_{\pi^{i}_{s}} (w^{i-1}_{s})\rangle
    \\
    &= \| v^{i-1}_s \|^2 +  \|f_{\pi^{i}_{s}} (w^{i}_{s}) - f_{\pi^{i}_{s}} (w^{i-1}_{s}) \|^2
    + \frac{2}{\eta}\langle w^{i-1}_s - w^i_s, f_{\pi^{i}_{s}} (w^{i}_{s}) - f_{\pi^{i}_{s}} (w^{i-1}_{s})\rangle.
\end{align*}
Assumption \ref{ass} (i) on convexity and $L$-smoothness of $f_{\pi^{i}_{s}}$ gives (see also Theorem 2.1.5 from \citep{nesterov2003introductory})
\begin{align*}
    \| v^{i}_s \|^2 &= \| v^{i-1}_s +  f_{\pi^{i}_{s}} (w^{i}_{s}) - f_{\pi^{i}_{s}} (w^{i-1}_{s}) \|^2
    \\
    &= \| v^{i-1}_s \|^2 +  \|f_{\pi^{i}_{s}} (w^{i}_{s}) - f_{\pi^{i}_{s}} (w^{i-1}_{s}) \|^2
    + 2\langle v^{i-1}_s, f_{\pi^{i}_{s}} (w^{i}_{s}) - f_{\pi^{i}_{s}} (w^{i-1}_{s})\rangle
    \\
    &= \| v^{i-1}_s \|^2 +  \|f_{\pi^{i}_{s}} (w^{i}_{s}) - f_{\pi^{i}_{s}} (w^{i-1}_{s}) \|^2
    + \frac{2}{\eta}\langle w^{i-1}_s - w^i_s, f_{\pi^{i}_{s}} (w^{i}_{s}) - f_{\pi^{i}_{s}} (w^{i-1}_{s})\rangle
    \\
    &\leq \| v^{i-1}_s \|^2 +  \left( 1 - \frac{2}{L \eta} \right)\|f_{\pi^{i}_{s}} (w^{i}_{s}) - f_{\pi^{i}_{s}} (w^{i-1}_{s}) \|^2.
\end{align*}
Taking into account that $\eta \leq \frac{1}{8nL} \leq  \frac{1}{2L}$, we finishes the proof.
\EndProof
\textbf{Proof of Theorem \ref{rr_th}}. For \texttt{RR-SARAH} $v_s = \nabla P(w_s)$, then by Lemma \ref{lemma2}, we get
\begin{align*}
     \left\| \nabla P(w_s) - \frac{1}{n+1} \sum\limits_{i=0}^{n} v^i_s\right\|^2 
     & \leq \left( \frac{4L^2}{n+1} + 4 \delta^2 n\right) \sum\limits_{i=1}^n \|w^{i}_s - w_s\|^2.
\end{align*}
Combining with Lemma \ref{lemma1}, one can obtain
\begin{align*}
    P(w_{s+1}) &\leq P(w_s) - \frac{\eta (n+1)}{2} \| \nabla P(w_s)\|^2 + \frac{\eta (n+1)}{2} \left( \frac{4L^2}{n+1} + 4 \delta^2 n\right) \sum\limits_{i=1}^n \|w^{i}_s - w_s\|^2.
\end{align*}
Next, we work with $\sum\limits_{i=1}^n \|w^{i}_s - w_s\|^2$. By Lemma \ref{lemma3} and the update for $w^i_s$ (\eqref{eq:w_0} and \eqref{eq:w_i_s}), we get
\begin{align}
    \label{eq:th2_temp1}
    \sum\limits_{i=1}^n \|w^{i}_s - w_s\|^2 &=  \eta^2 \sum\limits_{i=1}^n \left\| \sum\limits_{k=0}^{i-1} v^k_s \right\|^2 \leq \eta^2 \sum\limits_{i=1}^n i \sum\limits_{k=0}^{i-1} \left\|  v^k_s \right\|^2 \leq \eta^2 \sum\limits_{i=1}^n i \sum\limits_{k=0}^{i-1} \left\|  v_s \right\|^2 \notag\\
    &\leq \eta^2 \left\|  v_s \right\|^2 \sum\limits_{i=1}^n i \sum\limits_{k=0}^{i-1} 1
    \leq \eta^2 n^3 \left\|  v_s \right\|^2 = \eta^2 n^3 \left\|  \nabla P(w_s) \right\|^2. 
\end{align}
Hence,
\begin{align*}
    P(w_{s+1}) &\leq P(w_s) - \frac{\eta (n+1)}{2} \| \nabla P(w_s)\|^2 + \frac{\eta (n+1)}{2} \left( \frac{4L^2}{n+1} + 4 \delta^2 n\right) \cdot \eta^2 n^3 \left\| \nabla P(w_s)\right\|^2 \\
    &\leq P(w_s) - \frac{\eta (n+1)}{2} \left( 1 - \left( \frac{4L^2}{n+1} + 4 \delta^2 n\right) \cdot \eta^2 n^3\right) \| \nabla P(w_s)\|^2.
\end{align*}
With $\gamma \leq \min\left\{\frac{1}{8n L}; \frac{1}{8n^{2} \delta} \right\}$, we get
\begin{align*}
    P(w_{s+1}) - P^* &\leq P(w_s) - P^* - \frac{\eta (n+1)}{4} \| \nabla P(w_s)\|^2.
\end{align*}
Strong-convexity of $P$ ends the proof:
\begin{align*}
    P(w_{s+1}) - P^* &\leq \left(1 -  \frac{\eta (n+1) \mu}{2}\right)\left(P(w_s) - P^*\right).
\end{align*}
\EndProof
\textbf{Proof of Theorem \ref{th1}}. For \texttt{Shuffled-SARAH} $v_s = \frac{1}{n} \sum\limits_{i=1}^{n}  f_{\pi^{i}_{s-1}} (w^{i}_{s-1})$, then 
\begin{align}
    \label{eq:th1_temp1}
     &\left\| \nabla P(w_s) - \frac{1}{n} \sum\limits_{i=1}^{n} v^i_s\right\|^2 \notag\\
     &\hspace{1cm}\leq \left( \frac{4L^2}{n+1} + 4 \delta^2 n\right) \sum\limits_{i=1}^n \|w^{i}_s - w_s\|^2  + 2 \left\| \frac{1}{n} \sum\limits_{i=1}^{n}  \left[f_{\pi^{i}_{s-1}}(w_s) - f_{\pi^{i}_{s-1}} (w^{i}_{s-1}) \right] \right\|^2 \notag\\
     &\hspace{1cm}\leq \left(\frac{4L^2}{n+1} + 4\delta^2 n \right)  \sum\limits_{i=1}^n \|w^{i}_s - w_s\|^2  + \frac{2L^2}{n}\sum\limits_{i=1}^n  \left\| w^{i}_{s-1} - w_s\right\|^2.
\end{align}
With $\sum\limits_{i=1}^n \|w^{i}_s - w_s\|^2 $ we can work in the same way as in proof of Theorem \ref{rr_th}. In remains to deal with $\sum\limits_{i=1}^n  \left\| w^{i}_{s-1} - w_s\right\|^2$. Using Lemma \ref{lemma3} and the update for $w^i_s$ (\eqref{eq:w_0} and \eqref{eq:w_i_s}), we get
\begin{align}
\label{temp1}
    \sum\limits_{i=1}^n \|w^{i}_{s-1} - w_s\|^2 &=  \eta^2 \sum\limits_{i=1}^{n} \left\| \sum\limits_{k=1}^{n+ 1 -i} v^{n + 1-k}_{s-1} \right\|^2 \leq \eta^2 \sum\limits_{i=1}^n (n+1 -i) \sum\limits_{k=1}^{n+ 1 -i} \left\|  v^{n+1-k}_{s-1} \right\|^2 \nonumber\\
    &\leq \eta^2 \sum\limits_{i=1}^n (n+1 -i)\sum\limits_{k=1}^{n+ 1 -i}  \left\|  v_{s-1} \right\|^2 \nonumber\\
    &\leq \eta^2 \left\|  v_{s-1} \right\|^2 \sum\limits_{i=1}^n (n+1 -i) \sum\limits_{k=1}^{n+1 -i} 1  
    \nonumber\\
    & \leq \eta^2 n^3 \left\|  v_{s-1} \right\|^2. 
\end{align}
Combining the results of Lemma \ref{lemma1} with \eqref{eq:th1_temp1},\eqref{eq:th2_temp1} and \eqref{temp1}, one can obtain
\begin{align*}
    P(w_{s+1})
    \leq& P(w_s) - \frac{\eta (n+1)}{2} \| \nabla P(w_s)\|^2 \\
    &+ \frac{\eta (n+1)}{2} \left[ \left(\frac{4L^2}{n+1} + 4\delta^2 n \right) \cdot  \eta^2 n^3 \left\|  v_s\right\|^2  + \frac{2L^2}{n}\cdot \eta^2 n^3 \left\|  v_{s-1} \right\|^2\right] \\
    =& P(w_s) - \frac{\eta (n+1)}{4} \| \nabla P(w_s)\|^2 \\
    &+ \frac{\eta (n+1)}{2} \left[ \left(\frac{4L^2}{n+1} + 4\delta^2 n \right) \cdot  \eta^2 n^3 \left\|  v_s\right\|^2  + \frac{2L^2}{n}\cdot \eta^2 n^3 \left\|  v_{s-1} \right\|^2\right] \\
    &- \frac{\eta (n+1)}{4} \| \nabla P(w_s)\|^2 \\
    \leq& P(w_s) - \frac{\eta (n+1)}{4} \| \nabla P(w_s)\|^2 \\
    &+ \frac{\eta (n+1)}{2} \left[ \left(\frac{4L^2}{n+1} + 4\delta^2 n \right) \cdot  \eta^2 n^3 \left\|  v_s\right\|^2  + \frac{2L^2}{n}\cdot \eta^2 n^3 \left\|  v_{s-1} \right\|^2\right] \\
    &- \frac{\eta (n+1)}{8} \| v_s \|^2 + \frac{\eta (n+1)}{4} \| v_s - \nabla P(w_s)\|^2  \\
    \leq& P(w_s) - \frac{\eta (n+1)}{4} \| \nabla P(w_s)\|^2 \\
    &+ \frac{\eta (n+1)}{2} \left[ \left(\frac{4L^2}{n+1} + 4\delta^2 n \right) \cdot  \eta^2 n^3 \left\|  v_s\right\|^2  + \frac{2L^2}{n}\cdot \eta^2 n^3 \left\|  v_{s-1} \right\|^2\right] \\
    & - \frac{\eta (n+1)}{8} \| v_s \|^2 + \frac{\eta (n+1)}{4} \cdot \frac{2L^2}{n} \cdot \eta^2 n^3 \left\|  v_{s-1} \right\|^2.
\end{align*}
The last step is deduced the same way as \eqref{temp1}. Small rearrangement gives
\begin{align*}
    P(w_{s+1}) - P^*
    \leq& P(w_s) - P^* - \frac{\eta (n+1)}{4} \| \nabla P(w_s)\|^2 \\
    & - \frac{\eta (n+1)}{8} \left( 1 - \left(\frac{16L^2}{n+1} + 16\delta^2 n \right) \cdot  \eta^2 n^3 \right) \| v_s \|^2 \\
    &+ \eta (n+1)\cdot \frac{2L^2}{n} \cdot \eta^2 n^3 \left\|  v_{s-1} \right\|^2.
\end{align*}
With the choice of $\eta \leq \min\left\{\frac{1}{8n L}; \frac{1}{8n^{2} \delta} \right\}$, we have
\begin{align*}
    & P(w_{s+1}) - P^* + \frac{\eta (n+1)}{16}\| v_s \|^2 \\
    &\hspace{1cm} \leq P(w_s) - P^* - \frac{\eta (n+1)}{4} \| \nabla P(w_s)\|^2 + \frac{\eta (n+1)}{16}\cdot \frac{32L^2}{n} \cdot \eta^2 n^3 \left\|  v_{s-1} \right\|^2.
\end{align*}
Again using that $\eta \leq \frac{1}{8nL}$, we obtain $32L^2 \eta^2 n^2 \leq \left( 1 - \frac{\eta (n+1) \mu }{2}\right)$ and
\begin{align*}
    &P(w_{s+1}) - P^* + \frac{\eta (n+1)}{16}\| v_s \|^2 \\
    &\hspace{1cm}\leq P(w_s) - P^* - \frac{\eta (n+1)}{4} \| \nabla P(w_s)\|^2 + \left( 1 - \frac{\eta (n+1) \mu }{2}\right) \cdot  \frac{\eta (n+1)}{16}\left\|  v_{s-1} \right\|^2.
\end{align*}
Strong-convexity of $P$ ends the proof.
\EndProof

\end{document}